\documentclass{article} % For LaTeX2e
\usepackage[preprint]{neurips_2025}

\usepackage{amsmath,amsfonts,bm}

\def\eqref#1{equation~\ref{#1}}
\def\1{\bm{1}}

\DeclareMathAlphabet{\mathsfit}{\encodingdefault}{\sfdefault}{m}{sl}
\SetMathAlphabet{\mathsfit}{bold}{\encodingdefault}{\sfdefault}{bx}{n}

\usepackage{hyperref}
\usepackage{url}

\usepackage{graphicx}
\usepackage{subcaption}   % ✅ 必须包含这一句
\usepackage{float}        % 可选，用于强制图像位置
\usepackage{authblk}

\usepackage[utf8]{inputenc} % allow utf-8 input
\usepackage[T1]{fontenc}    % use 8-bit T1 fonts
\usepackage{hyperref}       % hyperlinks
\usepackage{url}            % simple URL typesetting
\usepackage{booktabs}       % professional-quality tables
\usepackage{amsfonts}       % blackboard math symbols
\usepackage{nicefrac}       % compact symbols for 1/2, etc.
\usepackage{microtype}      % microtypography
\usepackage{lipsum}		% Can be removed after putting your text content
\usepackage{graphicx}
\usepackage{natbib}
\usepackage{doi}
\usepackage{booktabs}
\usepackage{multirow}
\usepackage{appendix} 
\usepackage{float}
\usepackage{graphicx}
\usepackage{makecell}
\usepackage{subcaption}
\usepackage{xcolor}    
\usepackage{amsmath, array}
\usepackage{cleveref}       % for \cref cross-references

\usepackage{caption}
\usepackage{booktabs,tabularx,array}
\newcolumntype{L}{>{\raggedright\arraybackslash}X} % 左对齐的可换行列
\usepackage[ruled,vlined]{algorithm2e}
\newcommand{\C}{\mathcal{C}}
\renewcommand{\cite}[1]{\citep{#1}}

\title{In-Context Learning can Perform Continual Learning Like Humans}

\begin{document}

% Authors must not appear in the submitted version. They should be hidden
% as long as the \iclrfinalcopy macro remains commented out below.
% Non-anonymous submissions will be rejected without review.
\author[1]{Liuwang Kang}
\author[1]{\dag Fan Wang}
\author[1]{\dag Shaoshan Liu}
\author[1]{Hung-Chyun Chou}
\author[2]{Chuan Lin}
\author[1]{Ning Ding}

\affil[1]{Shenzhen Institute of Artificial Intelligence and Robotics for Society, Shenzhen, China}
\affil[2]{DeepRoute.ai, Shenzhen, China}

\maketitle

\def\thefootnote{\dag}\footnotetext{Corresponding to: fanwang.px@gmail.com, shaoshanliu@cuhk.edu.cn}

\begin{abstract}
Large language models (LLMs) can adapt to new tasks via in-context learning (ICL) without parameter updates, making them powerful learning engines for fast adaptation. While extensive research has examined ICL as a few-shot learner, whether it can achieve long-term retention and cross-task knowledge accumulation when multitasks arrive sequentially remains underexplored.
Motivated by human memory studies, we investigate the retention characteristics of ICL in multitask settings and extend it to in-context continual learning (ICCL), where continual learning ability emerges through task scheduling and prompt rearrangement.
Experiments on Markov-Chain benchmarks demonstrate that, for specific large-language models, ICCL benefits from distributed practice (DP) in a manner analogous to humans, consistently revealing a spacing ``sweet spot'' for retention. Beyond retention performance, we propose a human-retention similarity metric to quantify how closely a continual-learning (CL) method aligns with human retention dynamics. Using this metric, we show that linear-attention models such as MAMBA and RWKV exhibit particularly human-like retention patterns, despite their retention performance lagging behind that of Transformer-based LLMs.
Overall, our results establish ICCL as both cognitively plausible and practically effective, providing an inference-only CL paradigm that mitigates catastrophic forgetting and addresses the stability–plasticity dilemma in conventional CL methods.
\end{abstract}

\section{Introduction}
% \lipsum[2]
% \lipsum[3]

Large language models (LLMs) have demonstrated remarkable capabilities across a wide range of domains, with in-context learning (ICL) emerging as one of their most distinctive properties~\cite{brown2020language}. 
Via ICL, LLMs adapt new tasks on the fly by conditioning on demonstrations embedded in the prompt without parameter update.  This inference-only property turns them into versatile, efficient solvers that cope with zero-shot~\cite{ouyang2022training}, few-shot~\cite{brown2020language}, and many-shot~\cite{agarwal2024many} regimes, spanning Bayesian inference~\cite{xieexplanation}, regression~\cite{raventos2023pretraining,wu2023many}, and reinforcement learning~\cite{laskincontext,lee2023supervised,wang2025omnirl}. Such flexibility suggests that ICL could evolve into a general-purpose learning engine rather than remain a mere few-shot adapter~\cite{wang2025context}.
However, despite its promise, existing work seldom examines ICL’s ability to accumulate knowledge across tasks, confining it to one-off adaptation rather than a persistent learner.
Therefore, we raise the question of whether ICL can be extended toward continual learning, where knowledge can be maintained and reused across sequentially presented tasks.

In the broader literature, catastrophic forgetting (CF) has long been recognized as a fundamental challenge in gradient-based learning. To mitigate CF, a rich body of continual learning (CL) methods has been developed, including regularization-based approaches that reshape the objective function~\cite{li2020few,evilevitch2021avoiding}, replay-based approaches that reorganize the data stream~\cite{rolnick2019experience,buzzega2021rethinking}, and approaches that directly modify the optimization process~\cite{arous2021online,poggio2011online}. Collectively, these methods can be viewed as gradient-based continual learning (GBCL) since they predominantly rely on parameter updates through gradients. Although effective on many benchmarks, GBCL approaches fundamentally depend on parameter modification or explicit memory buffers and often struggle to achieve a desirable balance between stability and plasticity. This limitation motivates the exploration of a new paradigm: in-context continual learning (ICCL), which aims to extend ICL into continual learning scenarios by exploiting the model’s inherent capabilities rather than parameter updates.

\begin{table}[t]
\centering
\caption{Property comparisons among GBCL, ICL and ICCL.}
\label{tab:property_diff}
\resizebox{\linewidth}{!}{
\begin{tabularx}{\linewidth}{l L L L}
\toprule
\textbf{Property} &
\textbf{GBCL} &
\textbf{ICL} &
\textbf{ICCL} \\
\midrule
Parameter updates & Yes & No & No \\
Update rules & Gradient Descent & Blackbox Computation & Blackbox Computation \\
Knowledge carrier & Parameters & Memory/Hidden States &  Memory/Hidden States \\
Task settings & Multiple Tasks & Single Task & Multiple Tasks \\
\bottomrule
\end{tabularx}
}
\end{table}

Several recent studies have begun to explore directions related to ICCL, often by extending ICL with prompt modification (e.g., prompt tuning) or external augmentation (e.g., memory retrieval)~\cite{wang2022learning,gao2023retrieval,momeni2024context,shinwari2025memory}. While these approaches show promise, they also present several limitations. First, most efforts remain confined to single-task adaptation, without systematically addressing multitask sequential adaptation and cross-task knowledge accumulation. 
Second, they often rely on heuristic prompt engineering or auxiliary components that raise complexity yet remain vulnerable to memory bloat and retrieval errors, and plateau when procedural memory integration rather than fragmented retrieval is required.
Third, current evaluation benchmarks are often either too simple, resulting in trivial zero-shot solutions, or too complex, exceeding the capacity of LLMs to learn within context. As a result, they fail to adequately support systematic analyses of multitask learning dynamics. 

In this paper we focus on ICCL, which exploits the LLM’s built-in memory to flexibly control both learning and retention rather than external plug-and-play modules and manually specified learning rules, as summarized in Table~\ref{tab:property_diff}.
To compare GBCL and ICCL and to expose ICCL’s characteristics, we empirically examine the following perspectives:
First, in his pioneering studies of human memory, Hermann Ebbinghaus deliberately used “nonsense syllables” in serial-learning tasks~\cite{ebbinghaus2013image} to strip away prior knowledge and semantic associations, thereby isolating the core mechanisms of memory.
Guided by the same principle, we choose a benchmark for evaluating LLMs whose parameters already encode vast human knowledge: randomly generated Markov chains.
These chains preserve essential dynamical-system properties~\cite{attal2010markov,froyland2001extracting} while minimizing contamination from pre-existing knowledge.
Second, cognitive science shows that structured practice schedules enhance long-term memory of human~\cite{cepeda2006distributed,cepeda2008spacing,pavlik2005practice,bothell2020act}, underscoring the need to introduce task scheduling in machine continual learning.
Guided by this, we inject explicit task identifiers and cognition-inspired schedules to systematically probe how scheduling and prompts shape the retention characteristics of both LLMs and GBCL. Those includes single practice (SP), multiple practice (MP), and distributed practice (DP).
Experiments show that ICCL retains information more robustly than GBCL baselines and consistently privileges DP over SP and MP, mirroring cognitive findings on human memory.
Moreover, linear-attention models such as Mamba and RWKV exhibit retention dynamics that more closely parallel human behavior, even though their absolute retention lags behind standard self-attention.
In summary, our contributions are threefold:  
\begin{itemize}
    \item We investigate the retention characteristics of ICL in multitask settings and extend it to inference-only ICCL, where continual learning ability naturally emerges through task scheduling and prompt rearrangement.  
    \item We show that ICCL benefits from DP in a manner analogous to humans, consistently exhibiting a spacing sweet spot that enhances retention and provides a more favorable balance between stability and plasticity compared to GBCL baselines.
    \item We introduce a evaluation benchmark that combines retention performance with human-retention similarity, enabling systematic analysis of multitask retention and revealing that linear-attention models (e.g., MAMBA, RWKV) display particularly human-like retention patterns, despite lower absolute performance than Transformer-based LLMs.  
\end{itemize}

% The remainder of this paper is structured as follows. Section~\ref{sec:method} introduces details of methodology in ICCL and Section~\ref{sec:experiments} outlines experimental setup and reports the main results together with analyses. Section~\ref{sec:related_work} reviews a review of related work, highlighting their connections and differences with our work. Finally, Section~\ref{sec:conclusion} summarizes our findings.

\section{Methodology}
\label{sec:method}

\subsection{Formalization of ICCL}
\label{sec:iccl_define}

We now formalize the general protocol of ICCL. A task is denoted by $\tau: y^{\tau}_i \sim p_{\tau}(\cdot|x_i)$, which specifies a conditional distribution mapping an input $x_i$ to an output $y^{\tau}_i$. A segment of $\varphi$ experiences on task $\tau$ is represented as 
$\mathcal{D}_{\varphi}^{\tau}=\{x_1,y^{\tau}_1,\dots,x_{\varphi},y^{\tau}_{\varphi}\}$. 
To model ICCL, we consider a historical sequence of such experience segments $
\C_t=\bigoplus_{i=1}^{N}\mathcal{D}_{\varphi_i}^{\tau_i}$, 
where $t=\sum_{i=1}^{N}\varphi_i$ and $\oplus$ denotes ordered concatenation. Unlike a random collection, the order of segments is essential since ICCL is sensitive to the task order arrangement. 
Given a query input $x$ and the historical sequence $\C_t$, ICCL aims to approximate the output of an unknown target task $\tau^*$. Formally, we define:
\begin{equation}
\label{equ:iccl_define}
\hat{p}_{\theta}(y|x,\C_t) = \mathcal{F}_{\theta}\!\left(\bigoplus_{i=1}^{N}\mathcal{D}_{\varphi_i}^{\tau_i}, \, x\right) \;\Rightarrow\; p_{\tau^*}(y|x),
\end{equation}
where $\mathcal{F}_{\theta}$ denotes a sequential prediction structure parameterized by $\theta$, such as a large language model. Here, $p(y|x,\C_t)$ represents the probability of predicting output $y$ given $x$ and $\C_t$, and the goal is to approximate the true conditional distribution $p_{\tau^*}(y|x)$. Intuitively, ICCL leverages prior task experiences in context to infer the correct behavior on the target task.
Finally, tasks $\tau_i$ and $\tau^*$ are elements of a task collection $\mathcal{T}$, which defines the general scope of tasks considered in ICCL.

\textbf{Induction of task identifier.}  
Previous zero-shot instruction-following studies typically assume the availability of a task description $d^{\tau}$ that supplements the prompt~\cite{tanaka2024instructdoc,lou2023toward}. However, in practice, such meta-information is rarely known a priori and tasks must often be inferred implicitly. To accommodate this more realistic setting, we relax $d^{\tau}$ into a \emph{task identifier}, which need not encode semantic details but merely serves to distinguish one task from another. This abstraction aligns well with real-world applications: for instance, in preference modeling, the system may not have access to a user’s full profile but can instead rely on a user ID combined with the interaction history. Under this formulation, ICCL can be reformulated as
\begin{equation}
\label{equ:iccl_taskid}
\hat{p}_{\theta}(y|x,\C_t) = 
\mathcal{F}_{\theta}\!\Bigg(\bigoplus_{i=1}^{N}\big(d^{\tau_i}, \mathcal{D}_{\varphi_i}^{\tau_i}\big),\,d^{\tau^*},x\Bigg),
\end{equation}
where task identifiers $d^{\tau_i}$ and $d^{\tau^*}$ act as lightweight labels that allow the model to keep track of task boundaries without requiring explicit semantic descriptions.

\textbf{Evaluation of retention.}  
Given a historical sequence $\C_t$, we next evaluate how much information about the target task $\tau^*$ is retained after the model processes an intervening block $\mathcal{D}_{\varphi_D}^{\tau'}$ from another task $\tau' \neq \tau^*$. Specifically, this evaluation quantifies the extent to which the model "remembers" the target task $\tau^*$ after the $t$-step sequence $\C_t$ and an additional $\varphi_D$-step exposure to the distracting task $\tau'$ :
\begin{equation}
\label{equ:iccl_retention}
R_{\tau^*, \tau', \C_t, \theta}(t+\varphi_D) =
\mathbb{E}_{x,\tau'}\big[M\!\left(\hat{p}_{\theta}(\cdot|x,\C_t\oplus D^{\tau'}_{\varphi_D})),\;
p_{\tau^*}(\cdot|x)\right)\big],
\end{equation}
where $M(p_1, p_2)$ is a metric that quantifies the similarity between two probability distributions $p_1$ and $p_2$. We use \emph{Normalized Performance} (denoted by the symbol $R$) to assess the accuracy of $\hat{p}_{\theta}$, with $R(t) = 0$ indicating a complete loss of memory and $R(t) = 1$ indicating a perfect match between the estimator and the ground truth $p_{\tau^*}$.

\begin{figure}[h]
    \centering
    \includegraphics[width=0.85\linewidth]{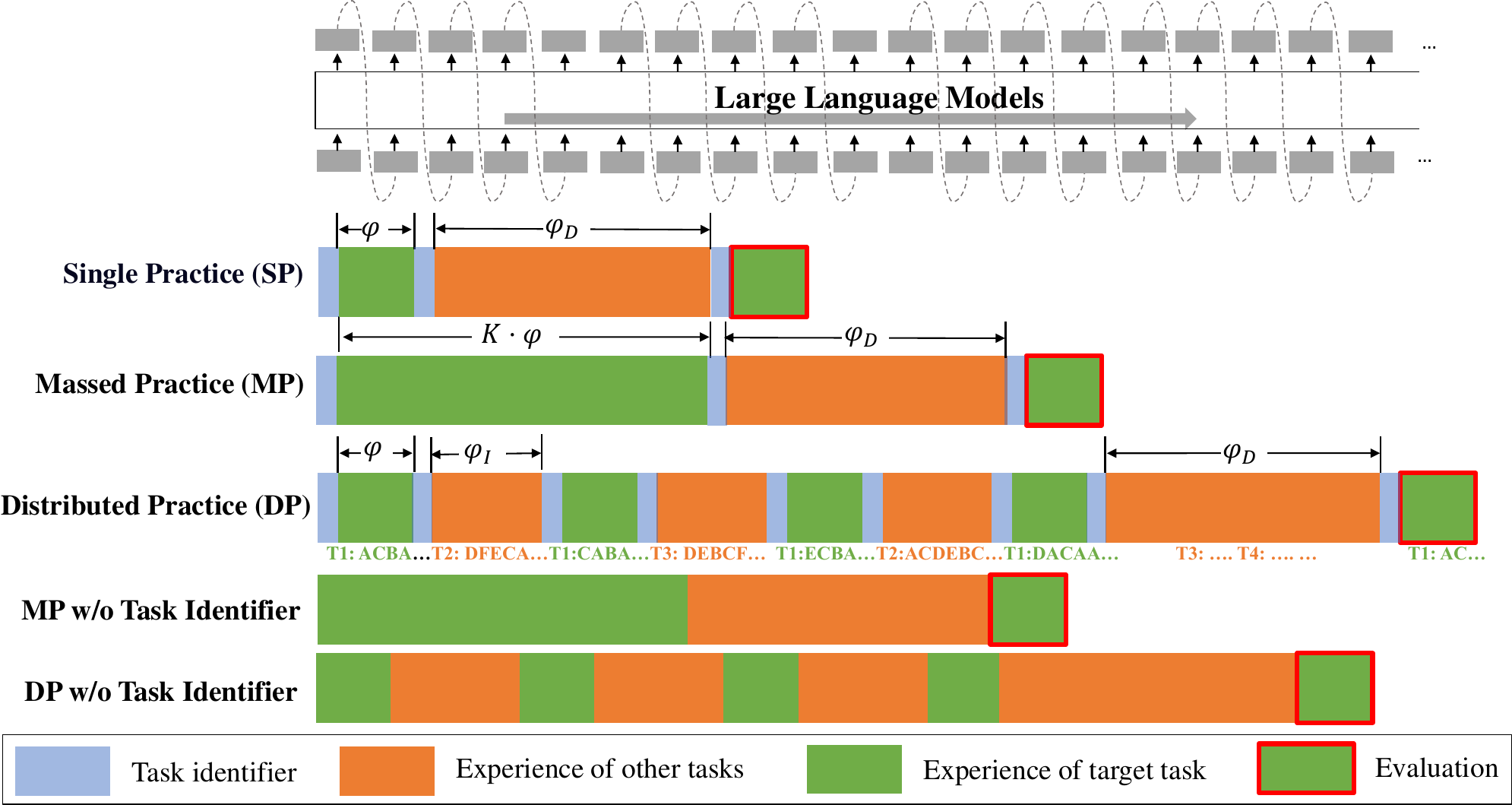}
    \caption{Illustration of how target and other task demonstrations are arranged under SP, MP, and DP, together with their variants without explicit task identifiers in a historical sequence.}
    \label{fig:evaluation_sketch}
    \vspace{-0.20in}
\end{figure}

\textbf{Scheduling strategy.} We formalize a scheduling strategy to capture how task demonstrations are arranged in a historical sequence $\C_t$. Specifically, different practice conditions correspond to different constructions of $\C_t$, as summarized as follows:

\begin{equation}
\label{equ:iccl_schedule}
\begin{aligned}
\text{Single Practice (SP):} \quad \C_t &= \mathcal{D}_{\varphi}^{\tau^*} \\
\text{Massed Practice (MP):} \quad \C_t &= \mathcal{D}_{K\varphi}^{\tau^*} \\
\text{Distributed Practice (DP):} \quad \C_t &= \bigoplus_{i=1}^{K}\!\big(\mathcal{D}_{\varphi}^{\tau^*},\mathcal{D}_{\varphi_{I}}^{\tau_i}\big), \quad \forall \tau_i \in \mathcal{T}, \; \tau_i \neq \tau^* .
\end{aligned}
\end{equation}

Eq.~(\ref{equ:iccl_schedule}) characterizes three canonical scheduling conditions, which are illustrated in Figure~\ref{fig:evaluation_sketch}.  
In \emph{SP}, demonstrations of the target task $\tau^*$ are presented only once in a contiguous block of length $\varphi$, after which the sequence moves to other tasks.  
In \emph{MP}, the demonstrations of the target task are extended to a larger block of length $K\varphi$, corresponding to the repeated rehearsal of $\tau^*$ without interruption.  
In \emph{DP}, the sequence alternates between the target task and other tasks: each target task block of length $\varphi$ is followed by another task block $\mathcal{D}_{\varphi_I}^{\tau_i}$ with $\tau_i \neq \tau^*$.  
This setup more closely resembles human retention patterns and allows ICCL to retain the target task knowledge in a manner analogous to human retention.  

To further investigate the role of explicit task identifiers, we additionally evaluate two variants where $d^{\tau}$ is removed from the prompt, denoted as \emph{MP w/o instruction} and \emph{DP w/o instruction}, respectively. 
These two variants test whether the presence of explicit task identifiers contributes to multitask sequential adaptation beyond the task order arrangement.

\subsection{Retention Dynamics in ICCL}
\label{sec:retention_dynamics}

A well-documented phenomenon in human retention is the spacing effect~\citep{anderson2004integrated,cepeda2008spacing}: MP triggers rapid decay, whereas DP yields superior long-term retention.  
To account for this effect, the ACT-R cognitive model~\citep{Borst2017ACTRFMRI,pavlik2005practice,petrov2006computationally} describes the resulting learning and retention curves.  
Suppose the learner practices (observes) the target task $\tau^*$ at the time stamp of $t_1,t_2,\dots,t_{K\varphi}$ and is exposed to other tasks otherwise; the ACT-R model predicts the probability of retaining the task at any subsequent time $t=t_{K\varphi} + \varphi_D$ by the following equation:
\begin{equation}
\hat{R}(t)=1/[1 + \exp(-\frac{w(t)-\gamma}{s})], \quad w(t)=\ln \sum_{i=1}^{K\varphi} [\kappa\cdot(t-t_i)]^{-d}
\label{eq_act_r}
\end{equation}
where $w(t)$ represents the memory activation that decays over time when no new practice of the target task occurs, and $P_{\tau^*}(t)$ denotes the probability of successfully accomplishing task $\tau^*$ at time $t$ given the memory activation $w(t)$. 
Notice that in our settings, for MP, $t_i=i$, for DP, $t_i=i + \left\lfloor \frac{i}{\varphi} \right\rfloor\varphi_I$.
The parameters $s$, $\gamma$, and $d$ are hyper-parameters to be determined. To bridge the gap between the actual time-dependency of human retention and the context length in ICCL, we introduce an additional hyper-parameter $\kappa$.

Our experimental results in Section~\ref{sec:sweet_spot} reveal an interior optimum of $\hat{R}(t)$ under DP, indicating the existence of a spacing sweet spot that maximizes the retention probability. This empirical pattern mirrors decades of findings in human memory research and provides practical guidance for prompt scheduling that avoids both massed and overly spaced practice.

\subsection{Quantifying Human Retention Similarity via Mahalanobis Distance}

We now ask a related question: \textit{to what extent do LLMs resemble human memory in their retention characteristics?} To enable a direct human--model comparison, we introduce Human Retention Similarity via Mahalanobis Distance (HRS-MD), which evaluates how closely the fitted ACT-R parameters of a model fall within parameter distributions reported in cognitive psychology. Specifically, Let the fitted parameter vector be
$\hat{\boldsymbol{\theta}} = [\hat{d},~\hat{s},~\hat{\gamma}]^{\top}$,
where $\hat{d}$ denotes the decay rate, $\hat{s}$ the activation noise, and $\hat{\gamma}$ the retrieval threshold estimated from retention curves. Human reference distributions are characterized by a mean vector
$\boldsymbol{\mu} = [\mu_d,~\mu_s,~\mu_\gamma]^{\top}$ and covariance matrix $\boldsymbol{\Sigma}$, derived from prior experimental studies~\cite{pavlik2005practice,LewisVasishth2005,bothell2020act,Borst2017ACTRFMRI,Said2016SugarFactory,VdVelde2022LBA}.  
HRS-MD is then defined as $(\hat{\boldsymbol{\theta}}-\boldsymbol{\mu})^{\top}
\boldsymbol{\Sigma}^{-1}
(\hat{\boldsymbol{\theta}}-\boldsymbol{\mu})$, which quantifies the discrepancy between the model-fitted parameters and human baselines while accounting for parameter scales and correlations. A smaller HRS-MD indicates closer alignment between LLMs and human memory retention. This metric enables systematic comparisons across ICCL and GBCL methods, revealing which model classes most closely reproduce human-like retention pattern. Details of $\boldsymbol{\Sigma}$ and reference distributions are provided in Appendix~\ref{appendix:mahalan_cal}.

\section{Experiment results}
\label{sec:experiments}

\subsection{Experiment Settings}

\paragraph{Models and Baselines.}
We evaluate two categories of methods: inference-only ICCL models and GBCL baselines. The ICCL group consists of four representative open-source LLMs: LLaMA3-8B, DeepSeek-R1, MAMBA, and RWKV-7. The GBCL group includes three classical approaches: Stochastic Gradient Descent (SGD)~\cite{arous2021online}, Experience Replay (ER)~\cite{rolnick2019experience}, and Elastic Weight Consolidation (EWC)~\cite{evilevitch2021avoiding}. 
For ICCL, experiences serve as the context in LLMs without updating any parameters; for GBCL, we construct a small predictive model that maps the input \& task-identifier to an output and sequentially refine it with the arriving experiences.
The detailed model architecture and hyperparameters are provided in the Appendix~\ref{appendix:model_struct}. 
To ensure fair comparisons, all models are evaluated on the same input sequences and prompt structures, with greedy decoding applied during inference. Each experiment is repeated 16 times with different random seeds, and results are reported as stable averages. 

\paragraph{Benchmark Task.}
Discrete Markov Chains (DMC) are frequently used benchmarking tasks~\cite{zekri2024large, akyurek2024context,liu2024llms}, as they capture essential properties of dynamical systems while minimizing contamination from pre-existing knowledge. It provides controlled task difficulty and a clean environment for analyzing retention dynamics. Two levels of complexity are considered: \textit{simple} tasks with $N_s=4$ states and \textit{complex} tasks with $N_s=8$ states. For each setting, random transition matrices are generated to sample states and construct experience sequences. Prompts are prepared under two conditions: (i) with task identifiers, where each sequence segment is prefixed with a label (e.g., \verb|[TARGET_TASK]|, \verb|[INTERFERENCE_TASK]|), and (ii) without identifiers, where the same content is presented unlabeled. This enables systematic evaluation of whether identifiers facilitate multitask sequential adaptation. Finally, we apply the scheduling strategies in Section~\ref{sec:iccl_define} to form prompts in SP, MP, and DP formats. Experimental parameters are set to $\varphi = 100$, $K = 5$, $\varphi_I \in {10, 50, 100, 200, 400, 600}$, and $\varphi_D \in (0,700)$.

\paragraph{Evaluation Metrics.}
A bottleneck in DMC evaluation is that performance divergence across tasks often stems from intrinsic task differences rather than prediction accuracy. To suppress this noise, we first introduce a normalized performance metric.
Inspired by the normalization schemes in~\cite{cha2007comprehensive}, we define the normalized retention performance metric $M$ (from \cref{equ:iccl_retention}) by rescaling the Bhattacharyya distance~\cite{liu2021makes, zekri2024large} $D_{Bah}(p_1, p_2)$ to the interval $[0,1]$:
\begin{equation}
M(\hat{p}_{\theta}, p_{\tau^*}) = 
\frac{\exp\!\big[D_{Bah}(p_{\tau^*},p_{\tau^*}) - D_{Bah}(\hat{p}_{\theta},p_{\tau^*})\big] - 1}
     {\exp\!\big[D_{Bah}(p_{\tau^*},p_{\tau^*}) - D_{Bah}(p_{\mathcal{U}},p_{\tau^*})\big] - 1},
\end{equation}
where $p_{\mathcal{U}}$ denotes the uniform distribution, which serves as the worst estimate of $p_{\tau^*}$.
All reported results are given as mean $\pm$ 95\% CI. In addition, we employ HRS-MD to measure how closely each method's fitted parameters align with human memory dynamics.

\subsection{Retention Performance Under SP, MP and DP}

\begin{figure}[h]
  \centering

  % ---------- Row 1 (3 figures) ----------
  \begin{subfigure}[b]{0.33\linewidth}
    \centering
    \includegraphics[width=\linewidth]{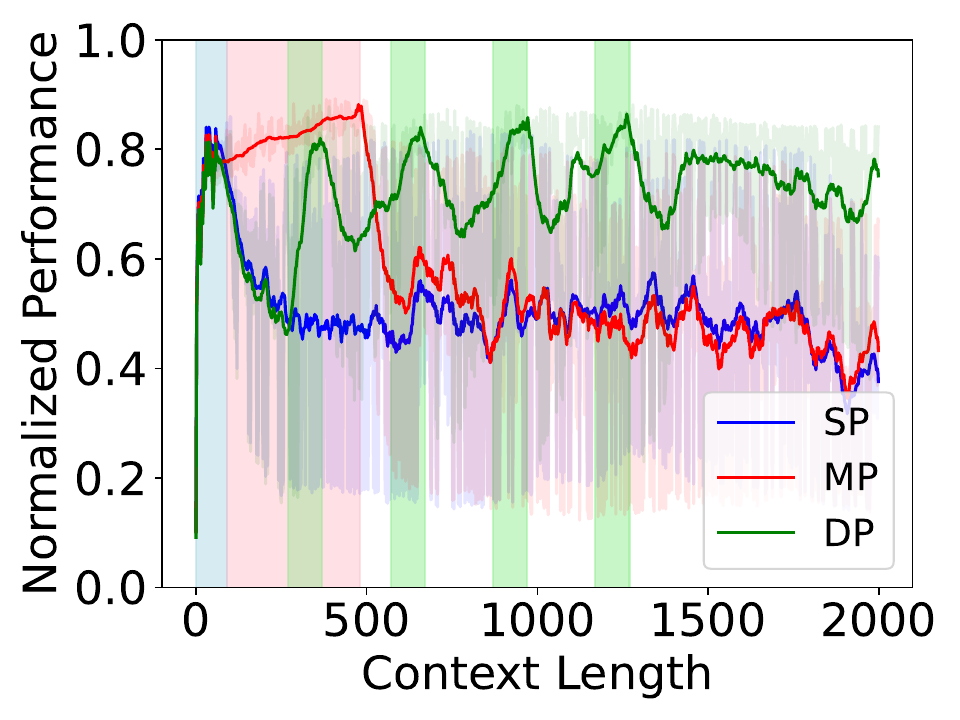}
    \caption{DEEPSEEK-R1}
    \label{fig:r1c1}
  \end{subfigure}\hfill
  \begin{subfigure}[b]{0.33\linewidth}
    \centering
    \includegraphics[width=\linewidth]{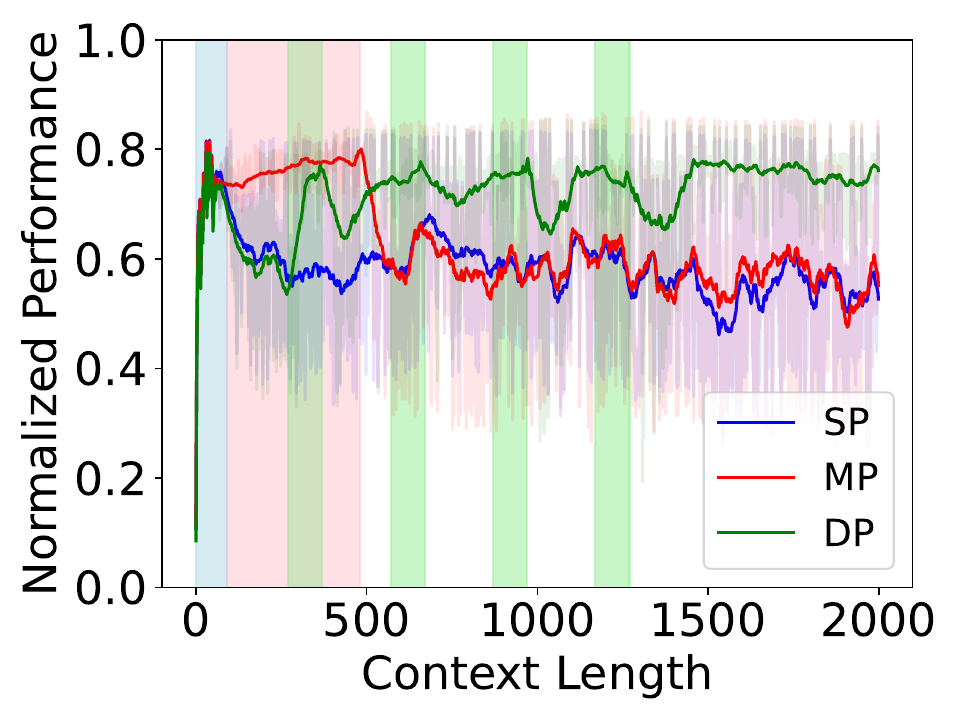}
    \caption{LLAMA3-8B}
    \label{fig:r1c2}
  \end{subfigure}\hfill
  \begin{subfigure}[b]{0.33\linewidth}
    \centering
    \includegraphics[width=\linewidth]{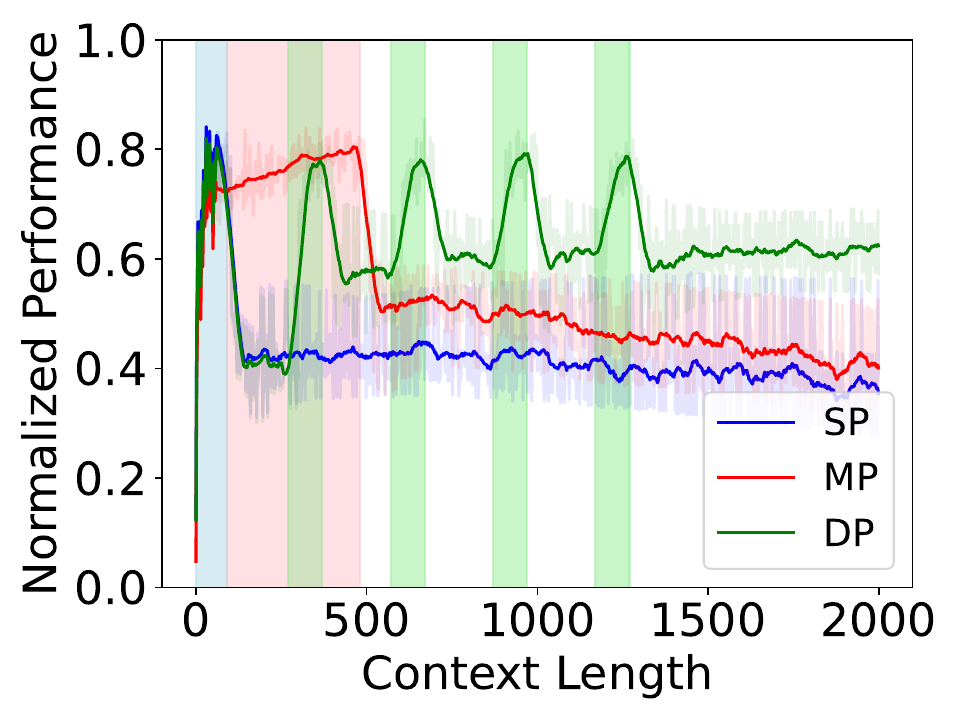}
    \caption{RWKV-7}
    \label{fig:r1c3}
  \end{subfigure}

  \vspace{0.6em}

  % ---------- Row 2 (3 figures) ----------
  \begin{subfigure}[b]{0.33\linewidth}
    \centering
    \includegraphics[width=\linewidth]{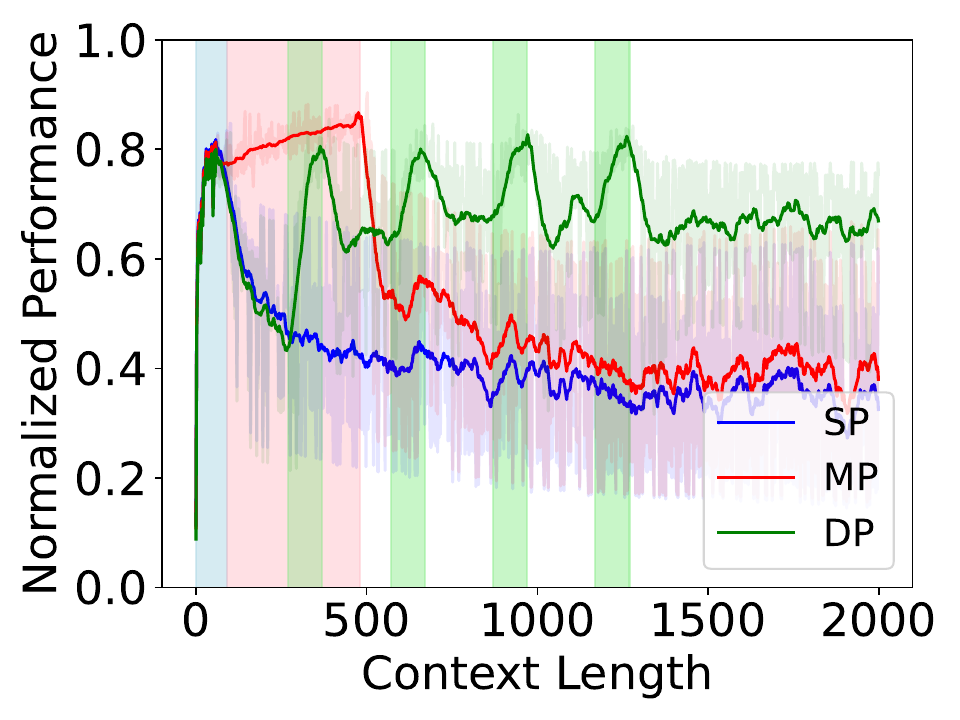}
    \caption{MAMBA}
    \label{fig:r2c1}
  \end{subfigure}\hfill
  \begin{subfigure}[b]{0.33\linewidth}
    \centering
    \includegraphics[width=\linewidth]{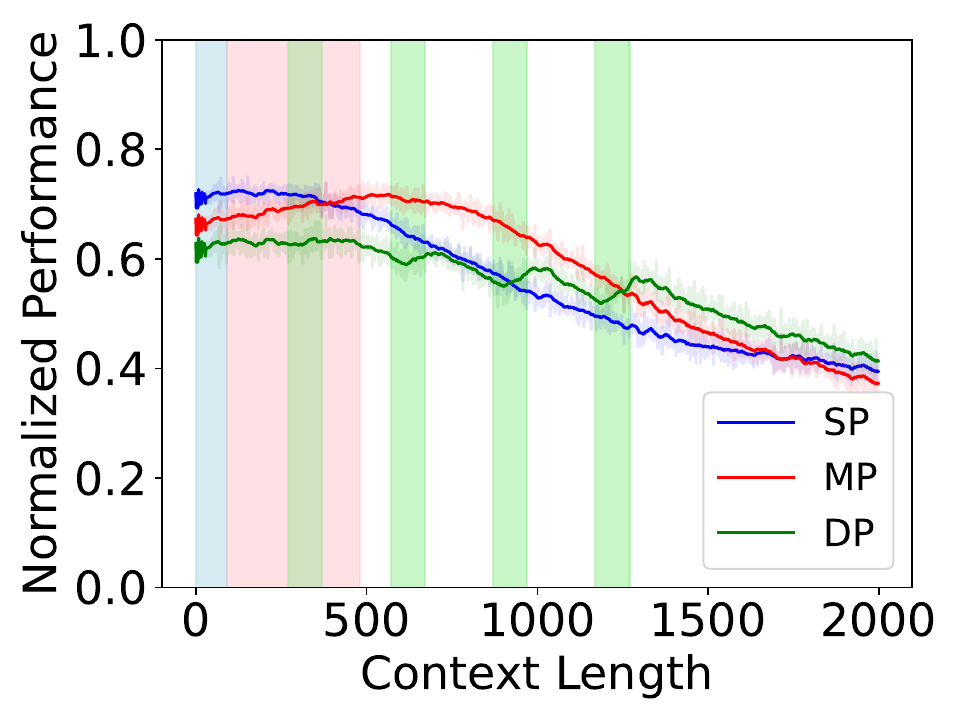}
    \caption{SGD}
    \label{fig:r2c2}
  \end{subfigure}\hfill
  \begin{subfigure}[b]{0.33\linewidth}
    \centering
    \includegraphics[width=\linewidth]{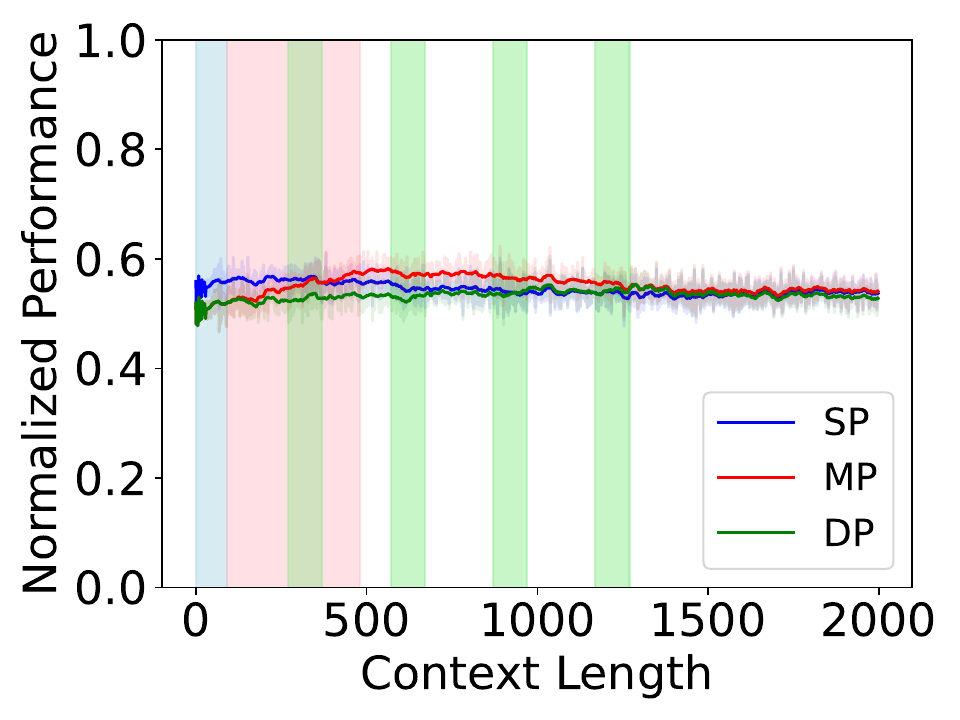}
    \caption{EWC}
    \label{fig:r2c3}
  \end{subfigure}
  \caption{Retention performance of ICCL and GBCL models on complex task under SP, MP, and DP settings (with task identifiers, $\varphi_I=200$). Target task blocks are highlighted in light blue (SP), pink (MP), and green (DP).}
  \label{fig:complex_task_curves}
\end{figure}

We first evaluate how the retention performance of ICCL models and GBCL baselines varies on simple and complex tasks with task identifiers and $\varphi_I=200$, under SP, MP, and DP scheduling. The experimental results on the complex task are shown in Figure~\ref{fig:complex_task_curves}. For both SP and MP, we observe that ICCL models exhibit a sharp performance drop immediately after the last target-task block, followed by a small continued decline as $\varphi_D$ increases, whereas some GBCL methods show a more gradual but consistent decrease--underscoring their susceptibility to catastrophic forgetting. Under DP, however, ICCL benefits substantially: DP effectively mitigates forgetting and maintains higher retention, while GBCL baselines show little improvement. Results for the ER baseline on the complex task and for the simple task are provided in Appendix~\ref{appendix:dp}.

Building on the above results, Table~\ref{tab:performance_interval} provides a detailed comparison of the retention performance of two randomly selected ICCL models and GBCL baselines under DP, evaluated on both simple and complex tasks as $\varphi_D$ increases from $0$ to $600$. The results show that ICCL models consistently outperform GBCL baselines. More importantly, although retention declines with larger $\varphi_D$, ICCL models sustain relatively high performance, whereas GBCL methods exhibit less stable outcomes across different $\varphi_D$ values.
In addition, increasing task complexity from simple to complex tasks reduces performance for both ICCL and GBCL. However, the degradation is markedly more severe for GBCL, underscoring that ICCL approaches not only achieve stronger overall retention under DP but also demonstrate greater robustness to task complexity.

\begin{table}[t]
\caption{Retention performance (mean ± 95\% CI) under DP for simple ($N_s=4$) and complex ($N_s=8$) Markov-chain tasks across varying intervals $\varphi_D$.}
\label{tab:performance_interval}
\resizebox{\linewidth}{!}{
\begin{tabular}{c|ccccc}
\hline
\multicolumn{6}{c}{\bfseries Simple Task (markov chain with $N_s =4$) } \\
\hline
\textbf{Model Types} & $\varphi_D=0$ & $\varphi_D=100$ & $\varphi_D=200$ & $\varphi_D=400$ & $\varphi_D=600$ \\
\hline
LLAMA3-8B & $\bf{0.819 \pm 0.086}$  & $\bf{0.737 \pm 0.090}$ & $\bf{0.865 \pm 0.030}$ & $\bf{0.833 \pm 0.067}$ & $\bf{0.781 \pm 0.078}$   \\
\hline
MAMBA & $0.668 \pm 0.106$ & $0.655 \pm 0.105$ & $0.678 \pm 0.094$ & $0.685 \pm 0.100$ & $0.667 \pm 0.101$   \\
\hline
SGD   & $0.606 \pm 0.100$ & $0.581 \pm 0.084$ & $0.558 \pm 0.086$ & $0.507 \pm 0.098$ & $0.471 \pm 0.101$   \\
\hline
ER    & $0.608 \pm 0.095$ & $0.613 \pm 0.075$ & $0.617 \pm 0.077$ & $0.620 \pm 0.081$ & $0.629 \pm 0.085$  \\
\hline
EWC    & $0.617 \pm 0.101$ & $0.612 \pm 0.086$ & $0.604 \pm 0.088$ & $0.601 \pm 0.099$ & $0.602 \pm 0.101$  \\
\hline
\multicolumn{6}{c}{\bfseries Complex Task (markov chain with $N_s =8$)} \\
\hline
\textbf{Model Types} & $\varphi_D=0$ & $\varphi_D=100$ & $\varphi_D=200$ & $\varphi_D=400$ & $\varphi_D=600$ \\
\hline
LLAMA3-8B & $\bf{0.784 \pm 0.054}$ & $\bf{0.776 \pm 0.056}$ & $\bf{0.753 \pm 0.086}$ & $\bf{0.783 \pm 0.057}$ & $\bf{0.792 \pm 0.058}$  \\
\hline
MAMBA    & $0.750 \pm 0.059$ & $0.700 \pm 0.061$ & $0.608 \pm 0.103$ & $0.724 \pm 0.064$ & $0.726 \pm 0.048$  \\
\hline
SGD   & $0.577 \pm 0.072$ & $0.540 \pm 0.076$ & $0.491 \pm 0.080$ & $0.464 \pm 0.073$ & $0.448 \pm 0.085$   \\
\hline
ER    & $0.519 \pm 0.081$ & $0.501 \pm 0.086$ & $0.472 \pm 0.088$ & $0.479 \pm 0.080$ & $0.495 \pm 0.091$  \\
\hline
EWC    & $0.563 \pm 0.079$ & $0.551 \pm 0.087$ & $0.526 \pm 0.091$ & $0.537 \pm 0.083$ & $0.542 \pm 0.093$  \\
\hline
\end{tabular}
}
\vspace{-0.5em} 
\end{table}

\subsection{Effect of Task Identifier on ICCL Retention}

\begin{figure}[h]
    \centering
    \begin{subfigure}[b]{0.48\textwidth}
        \centering
        \includegraphics[width=\linewidth]{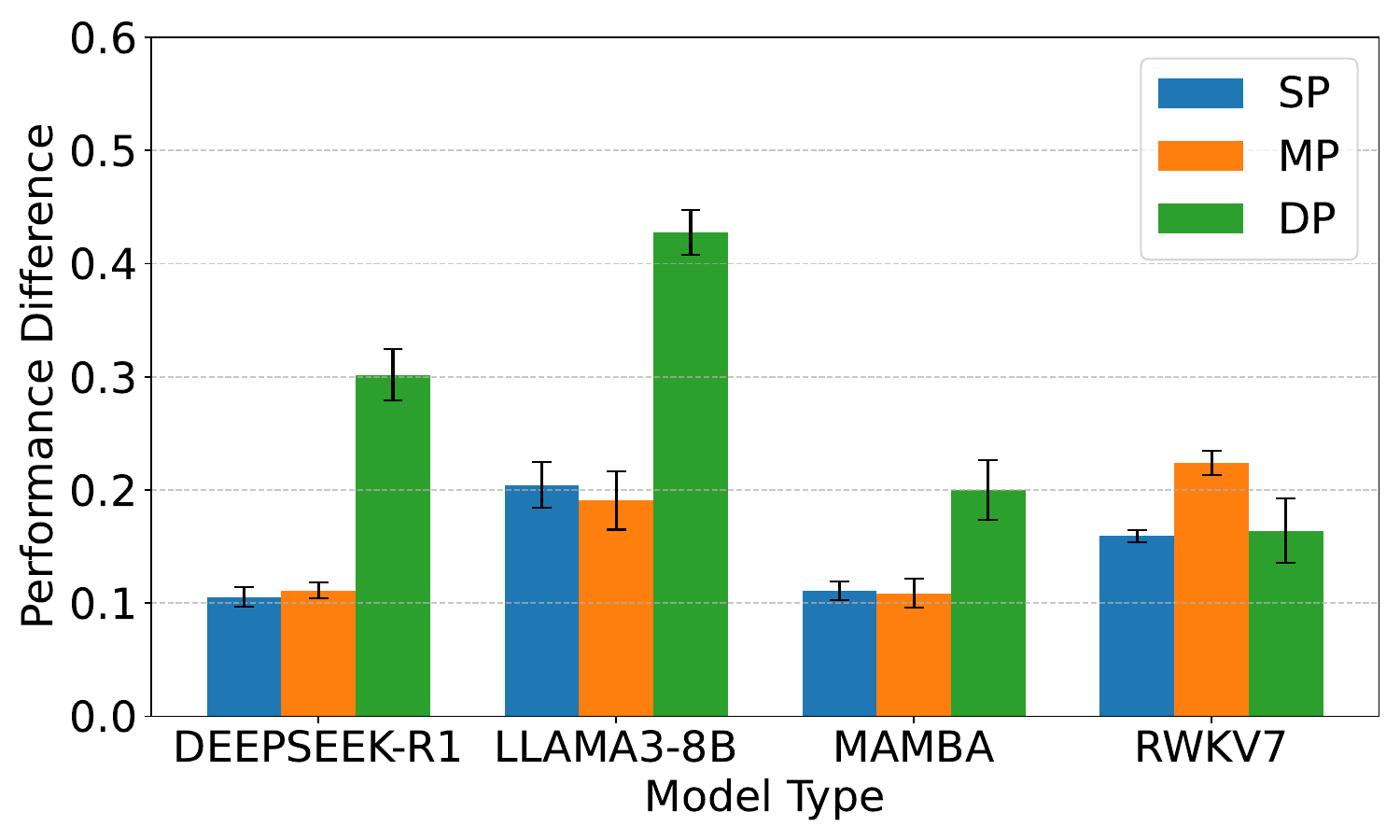}
        \caption{Simple Task (markov chain with $N_s = 4$)}
        \label{fig:model1}
    \end{subfigure}
    \hfill
    \begin{subfigure}[b]{0.48\textwidth}
        \centering
        \includegraphics[width=\linewidth]{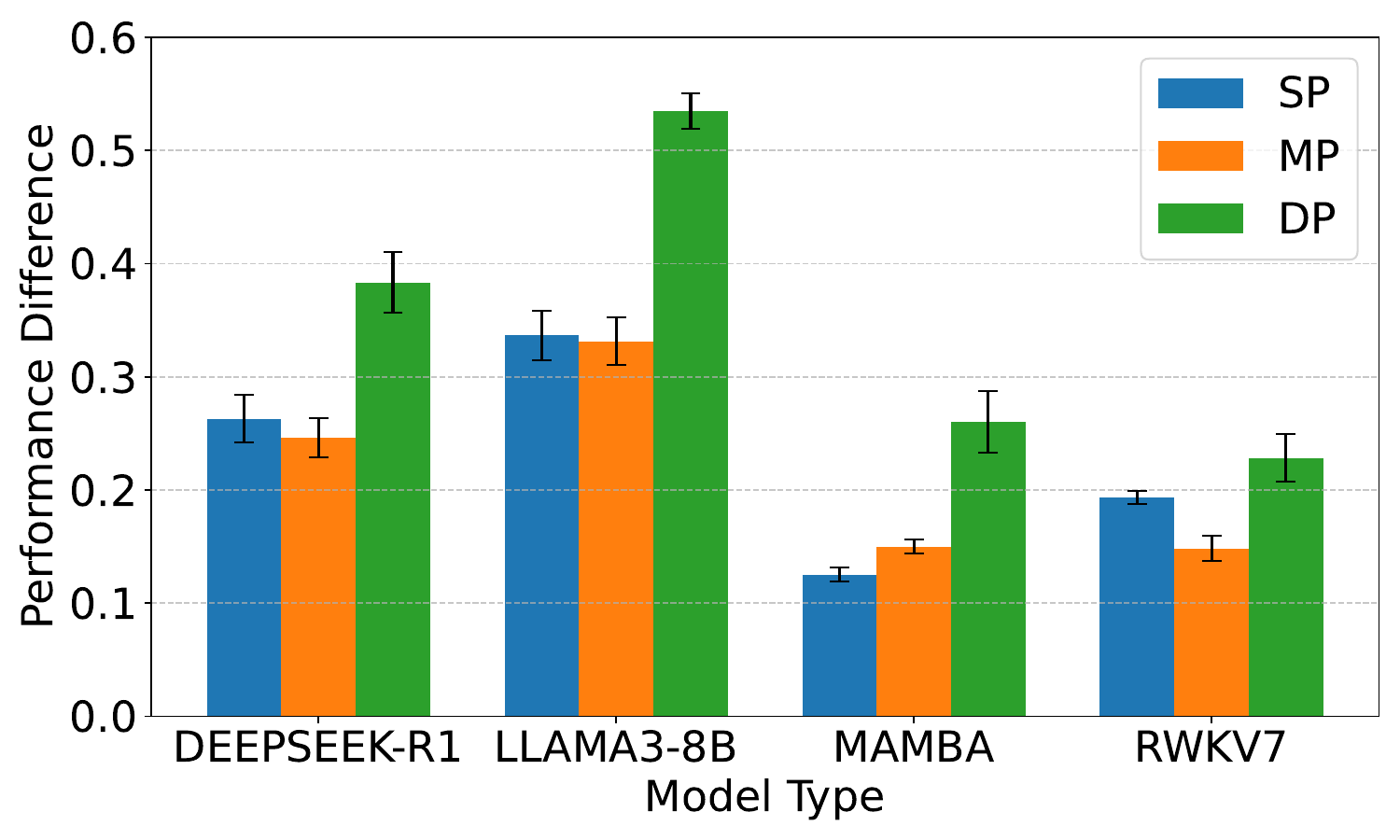}
        \caption{Complex Task (markov chain with $N_s = 8$)}
        \label{fig:model2}
    \end{subfigure}
    
    \caption{Retention performance differences between prompts with and without task identifiers across ICCL models under SP, MP, and DP ($\varphi_I=200$).}
    \label{fig:four_models_task_identifier}
\end{figure}

Figure~\ref{fig:four_models_task_identifier} compares the retention performance difference of ICCL models with and without task identifiers under SP, MP, and DP ($\varphi_I=200$). The results show that incorporating task identifiers consistently enhances performance across all models and scheduling strategies, with the largest gains observed under DP. Moreover, while task identifiers improve ICCL retention on both simple and complex tasks, the performance gap is noticeably larger in the complex setting. This suggests that explicit identifiers become increasingly important as task complexity grows. Overall, these findings underscore the critical role of task identifiers in strengthening ICCL retention, particularly in complex multitask scenarios.

\subsection{Spacing Sweet Spot in ICCL under DP}
\label{sec:sweet_spot}

\begin{figure}[h]
    \centering
    \begin{subfigure}[b]{0.98\textwidth}
        \centering
        \includegraphics[width=\linewidth]{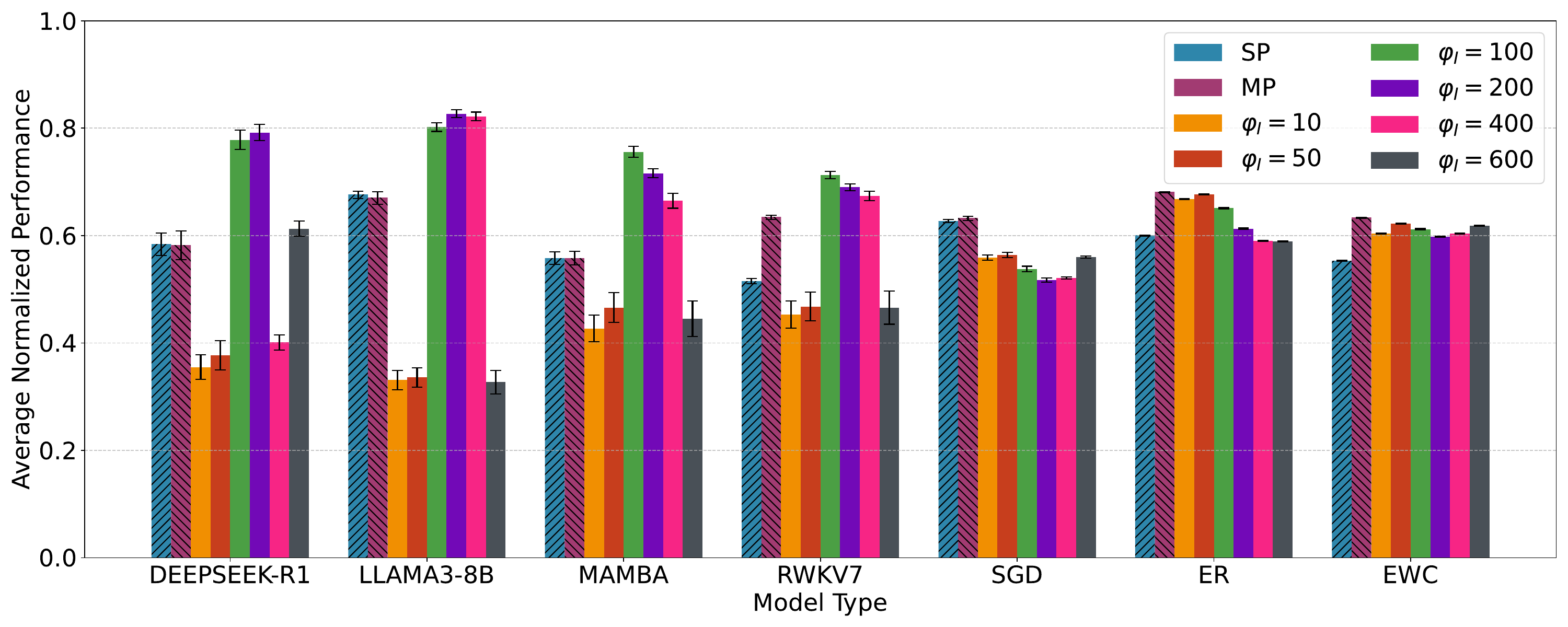}
        \caption{Simple Task (markov chain with $N_s =4$)}
        \label{fig:model1}
    \end{subfigure}
    \hfill
    \begin{subfigure}[b]{0.98\textwidth}
        \centering
        \includegraphics[width=\linewidth]{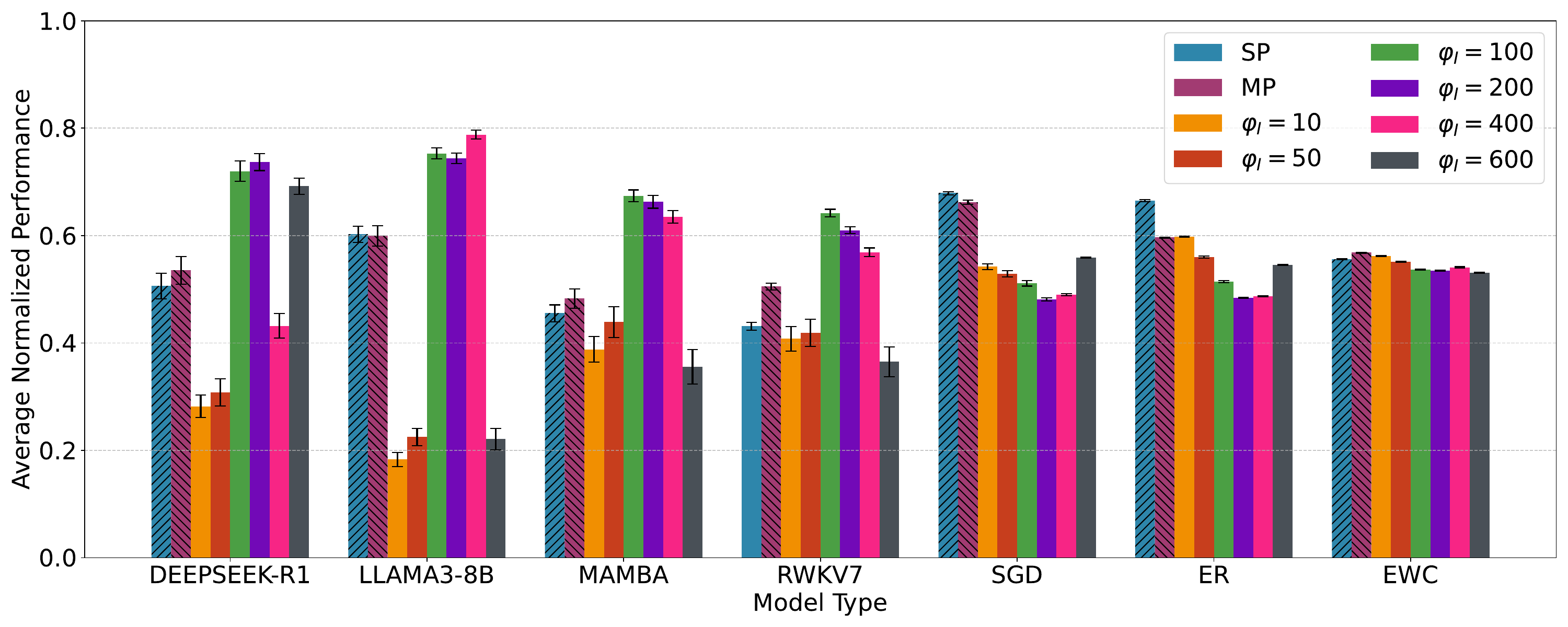}
        \caption{Complex Task (markov chain with $N_s =8$)}
        \label{fig:model2}
    \end{subfigure}
    
    \caption{Average retention performance comparison among ICCL and GBCL methods under DP with different intervals ($\varphi_I$) on both simple and complex tasks.}
    \label{fig:seven_models_dp}
\end{figure}

Figure~\ref{fig:seven_models_dp} reports the average retention performance of ICCL and GBCL models under DP across different intervals $\varphi_I$, evaluated on both simple and complex tasks. Here, average retention performance is defined as the mean retention over $\varphi_D$, thereby capturing memory retention beyond the immediate target-task exposure.
The results reveal a clear contrast between ICCL and GBCL. For ICCL, DP induces a characteristic \emph{spacing sweet spot}, where average retention performance peaks at an intermediate $\varphi_I$ lying between 100 and 400. This sweet spot consistently emerges across both simple and complex tasks, although average retention performance is higher in the simpler setting. These findings empirically support the presence of a spacing sweet spot and further demonstrate that ICCL under DP surpasses the performance achieved with SP and MP. Thus, ICCL benefits from DP in a manner analogous to human memory, and careful tuning of $\varphi_I$ can substantially enhance retention. In contrast, GBCL baselines exhibit no such sweet spot: their performance remains largely flat or gradually declines as $\varphi_I$ increases, indicating that DP provides little to no advantage.

\subsection{HRS-MD Quantification}

\begin{table}[t]
\centering
\caption{Fitted ACT-R parameters with corresponding MSE and HRS-MD values.}
\label{tab:actr_params}
\resizebox{\linewidth}{!}{
\begin{tabular}{lcccccc}
\toprule
\textbf{Model} & \makecell{\textbf{Decay Rate}\\($d$)} & \makecell{\textbf{Activation Noise}\\($s$)} & \makecell{\textbf{Scaling Factor}\\($\kappa$)} & \makecell{\textbf{Threshold}\\($\gamma$)} & \textbf{MSE} & \textbf{HRS-MD}\\
\midrule
LLAMA3      & 0.14 & 2.00 & 1.20 & 1.01 & 0.0024 & 500.22 \\
DEEPSEEK-R1 & 0.35 & 1.69 & 0.43 & -0.24 & 0.0155 & 302.39 \\
RWKV-7      & 0.27 & 1.62 & 0.94 & 0.59 & 0.0052 & \textbf{287.57} \\
MAMBA       & 0.29 & 1.59 & 0.76 & 0.33 & 0.0049 & \textbf{272.74} \\
SGD         & 0.41 & 2.00 & 0.77 & 0.15 & 0.0036 & 445.07 \\
ER          & 0.27 & 2.00 & 1.36 & 1.02 & 0.0061 & 466.74 \\
EWC         & 0.22 & 2.00 & 1.31 & 1.65 & 0.0021 & 481.53 \\
\bottomrule
\end{tabular}
}
\vspace{-0.5em} 
\end{table}

To investigate how the aforementioned retention curves align with human retention, we employed the human-retention-validated ACT-R model to fit the ICCL and GBCL curves. Specifically, we used \cref{eq_act_r} to model the normalized ICCL/GBCL performance obtained from \cref{equ:iccl_retention}.
Table~\ref{tab:actr_params} summarizes the fitted ACT-R parameters, mean squared error (MSE) during parameter fitting, and HRS-MD values for both ICCL and GBCL methods based on retention performance results on both simple and complex tasks under DP across different intervals $\varphi_I$. Several consistent patterns emerge.
First, ICCL models--except for LLaMA3--achieve lower HRS-MD values than GBCL baselines, indicating that their retention dynamics more closely approximate human memory. In particular, MAMBA and RWKV-7 yield the lowest HRS-MD scores ($272.74$ and $287.57$, respectively), demonstrating the strongest alignment with human retention patterns. By contrast, GBCL methods produce substantially higher HRS-MD values, reflecting greater divergence from human memory dynamics.
Second, the fitted parameters reveal distinctive tendencies across the two paradigms. ICCL models generally converge to moderate decay rates, relatively low activation noise, and thresholds near zero—indicating efficient and accessible retrieval dynamics, particularly evident in MAMBA and RWKV-7. In contrast, GBCL methods exhibit more extreme parameter configurations: SGD yields the highest decay rate ($d=0.41$), consistent with rapid forgetting, while ER and EWC converge to elevated thresholds ($\gamma \geq 1.0$), implying stricter retrieval conditions and reduced accessibility of stored knowledge. These patterns highlight the poor balance between stability and plasticity in GBCL approaches.

\section{Related Work}
\label{sec:related_work}

\textbf{GBCL} represents the classical paradigm for mitigating catastrophic forgetting in sequential learning tasks. By updating model parameters, GBCL aims to balance plasticity with stability. Representative approaches fall into three categories: EWC, ER, and SGD. EWC constrains updates to parameters deemed important for past tasks. For instance, work~\cite{li2020few,evilevitch2021avoiding} penalizes deviations of critical parameters using fisher-information-based importance scores, while synaptic intelligence~\cite{zenke2017continual,zenkeimproved} accumulates importance online by measuring each parameter’s contribution to past loss reductions. ER preserves a small memory of past examples and mixes them with new inputs, as in experience replay~\cite{rolnick2019experience,buzzega2021rethinking}. Finally, SGD~\cite{arous2021online,poggio2011online} updates parameters in real time using only current data, ensuring adaptability but suffering severe forgetting when task distributions shift. In all cases, GBCL requires parameter updates or memory buffers to maintain adaptation and has poor balance between stability and plasticity.  

\textbf{ICL} is an emergent capability of LLMs that enables rapid task adaptation through demonstrations in prompts, without parameter updates. Prior work has studied ICL empirically examining sensitivity to prompt tuning, label distribution, and demonstration ordering~\cite{zhao2021calibrate,liu2021makes,lu2021fantastically,liu2023lost} and theoretically, framing it as implicit Bayesian inference~\cite{xie2021explanation}, gradient-descent-like adaptation in attention~\cite{akyurek2022learning,von2023transformers}, or kernel regression over in-context samples~\cite{han2023explaining}. Scaling studies further show that many-shot ICL extends to regression and decision-tree tasks~\cite{agarwal2024many,garg2022can}. However, most research efforts remain confined to single-task adaptation, without systematically addressing multitask sequential adaptation and cross-task knowledge accumulation. Recent studies attempt to extend ICL toward continual learning, either by incrementally adapting prompts~\cite{wang2022learning} or by augmenting them with external retrieval mechanisms~\cite{gao2023retrieval,momeni2024context,shinwari2025memory}. Despite these advances, existing approaches often rely on heuristic prompt engineering or auxiliary components, which add complexity but remain vulnerable to memory bloat and retrieval errors. 
% By contrast, ICCL leverages cognitively inspired scheduling strategies and benchmarks to better balance stability and plasticity.  

\textbf{Human memory research} provides systematic insights into retention dynamics. Ebbinghaus~\cite{ebbinghaus2013image} first introduced the forgetting curve, showing exponential-like memory decay, later refined into power-law models~\cite{white2001forgetting} that more accurately capture systematic forgetting. Building on this foundation, the \emph{spacing effect} demonstrated that distributed practice yields substantially better long-term retention than massed practice. Large-scale meta-analyses~\cite{cepeda2006distributed,cepeda2008spacing} confirmed the robustness of this effect and identified optimal inter-study intervals. To model these dynamics, Anderson~\cite{pavlik2005practice} proposed the base-level learning equation linking recall probability to frequency and recency of exposures, and Pavlik~\cite{pavlik2008using} extended it into an ACT-R framework for computing optimal practice schedules~\cite{bothell2020act,Borst2017ACTRFMRI}. 
% While originally derived from human cognition, these models suggest generalizable memory principles. 
Our work builds on this link by testing whether LLMs exhibit analogous human-like retention pattern when demonstrations are distributed in prompts.

\section{Conclusion}
\label{sec:conclusion}

In this work, we investigated the retention characteristics of ICL in multi-task settings and extended it to ICCL through prompt scheduling and rearrangement. Experiments on Markov-chain benchmarks showed that ICCL benefits DP in a manner analogous to humans, consistently revealing a spacing sweet spot that enhances retention beyond SP and MP settings. Beyond retention performance, we introduced the HRS-MD metric to quantify alignment with human memory dynamics, and found that linear-attention models such as MAMBA and RWKV display particularly human-like retention patterns. Overall, our results establish ICCL as both cognitively plausible and practically effective, providing an inference-only paradigm that mitigates catastrophic forgetting and addresses the stability–plasticity dilemma in conventional continual learning methods.

\begin{ack}
This project is supported by Longgang District Shenzhen's “Ten Action Plan” for supporting innovation projects (under Grant  LGKCSDPT2024002).
\end{ack}

\bibliography{arxiv}
\bibliographystyle{unsrtnat}

\appendix
\section{Appendix}

\subsection{Use of LLMs}
We used large language models (LLMs) only as an auxiliary tool to improve the clarity and presentation of this paper. 
The assistance was limited to:
\begin{itemize}
    \item \textbf{Language refinement:} grammar checking, wording suggestions, and improving sentence fluency while preserving the authors’ original technical content.
    \item \textbf{Mathematical support:} verifying the correctness and readability of some derivations and notations, without introducing new technical results.
\end{itemize}

No LLM was used for generating research ideas, designing experiments, analyzing results, or writing original scientific content. 
All conceptual and technical contributions were made by the authors. 
The authors take full responsibility for the entire content of this paper, and LLMs are not eligible for authorship.

\subsection{Human Reference Distribution and Covariance in Mahalanobis Distance}
\label{appendix:mahalan_cal}

We evaluate the plausibility of the fitted ACT-R model parameters
\(
\hat{\boldsymbol{\theta}}
=
[\hat d,\,\hat s,\,\hat\gamma]^\top
\)
with respect to a human reference distribution
characterized by a mean vector
\(
\boldsymbol{\mu}=[\mu_d,\,\mu_s,\,\mu_\gamma]^\top
\)
and a covariance matrix
\(
\boldsymbol{\Sigma}\).
Given \(\boldsymbol{\mu}\) and \(\boldsymbol{\Sigma}\),
the squared Mahalanobis distance is
\[
D^2
=
(\hat{\boldsymbol{\theta}}-\boldsymbol{\mu})^{\top}
\boldsymbol{\Sigma}^{-1}
(\hat{\boldsymbol{\theta}}-\boldsymbol{\mu}),
\]
which normalizes deviations by considering parameter scales and correlations. We report an interpretable similarity score \(\mathrm{Score}=\exp(-\tfrac{1}{2}D^2)\), interpretable as a Gaussian RBF under the Mahalanobis metric, where larger values indicate closer alignment with the human reference.

\begin{table}[t]
\caption{Statistical reference means and standard derivations for human's ACT-R model parameters according to~\cite{pavlik2005practice,LewisVasishth2005,bothell2020act,Borst2017ACTRFMRI,Said2016SugarFactory,VdVelde2022LBA}.}
\label{tab:empical_parameters}
\centering
\begin{tabular}{lccc|}
\hline
\textbf{Parameter} & Mean value ($\mu$) & Standard derivation ($\sigma$) \\
\hline
Decay Rate ($d$) & \(\mu_d = 0.50\) & \(\sigma_d = 0.05\) \\
Activation Noise ($s$) & \(\mu_s = 0.32\) & \(\sigma_s = 0.08\) \\
Threshold ($\gamma$) & \(\mu_{\gamma} = -0.50\) & \(\sigma_{\gamma} = 0.71\) \\
\hline
\end{tabular}
\end{table}

\paragraph{Estimating the covariance matrix \(\boldsymbol{\Sigma}\).}
Table~\ref{tab:empical_parameters} reports the reference means and standard deviations of human ACT-R parameters from prior studies. Since only marginal statistics are available and raw human samples (hence empirical correlations) are not, we construct the covariance matrix from these reported scales. Let
\(\boldsymbol{\sigma}=[\sigma_d,\,\sigma_s,\,\sigma_\gamma]^\top\) denote the standard deviations and \(\mathbf{D}=\mathrm{diag}(\boldsymbol{\sigma})\).

By default, we adopt the independence approximation, assuming zero cross-covariances:
\[
\boldsymbol{\Sigma}
=\mathrm{diag}\!\big(\sigma_d^2,\,\sigma_s^2,\,\sigma_\gamma^2\big).
\]
This choice reflects the fact that no reliable correlation estimates are available and reduces the Mahalanobis distance to a multidimensional $z$-score distance.

\subsection{Model architecture and parameters}
\label{appendix:model_struct}

\paragraph{GBCL.}
SGD, ER, and EWC employ the neural network structure. 
The neural network takes as input the current discrete state and a task identifier. By default, both are embedded through separate layers of size $\tfrac{d}{2}$ (with $d=64$), concatenated, and passed to a multilayer perceptron with two hidden layers and ReLU activations:
The output produces logits over the next-state space, which are transformed into probabilities via softmax. 
Weights of linear and embedding layers are initialized with Xavier uniform initialization, and biases are set to zero. 
Unless otherwise specified, we set $d=64$, \texttt{num\_states}$=4/8$, and \texttt{num\_tasks}$=2$. 
Optimization in all cases uses stochastic gradient descent with cross-entropy loss. 
While the network architecture is identical across methods, their training hyperparameters differ. 
For SGD, the model is updated online using only the current sample with a learning rate of $0.001$. In ER, a replay buffer is employed with a capacity of $8000$ samples, a replay ratio of $0.5$, and a batch size of $32$. 
For EWC), the model is trained with the same online updates as SGD but augmented with a quadratic penalty weighted by the diagonal Fisher information matrix; the optimal regularization strength is set to $\lambda_{\text{EWC}}=700$.

\paragraph{LLMs.}
LLaMA3-8B~\cite{llama3_8b_instruct}, DeepSeek-R1~\cite{deepseek_coder_7b}, MAMBA~\cite{falcon_mamba_7b}, and RWKV-7~\cite{rwkv_4_world_7b} are used in their publicly released configurations without further fine-tuning. 
LLaMA3-8B is a transformer-based decoder model with approximately 8 billion parameters and a context length of 8K tokens.  
DeepSeek-R1 is a transformer model with roughly 7 billion parameters, optimized for reasoning tasks, supporting up to 8K tokens.  
MAMBA is a selective state-space model with about 8 billion parameters and an extended context length of 16K tokens.  
RWKV-7 adopts a hybrid RNN–transformer architecture, with around 7 billion parameters and a context length of 4K tokens.  
For all large models, we adopt the default inference hyperparameters recommended by their authors: greedy or temperature-controlled decoding with temperature in $[0.6,0.8]$, maximum generation length matched to the context window, and no additional training beyond the released checkpoints.

\subsection{Retention Performance under SP, DP and MP for Markvo Chains}
\label{appendix:dp}

\begin{figure}[h]
  \centering

  % ---------- Row 1 (3 figures) ----------
  \begin{subfigure}[b]{0.33\linewidth}
    \centering
    \includegraphics[width=\linewidth]{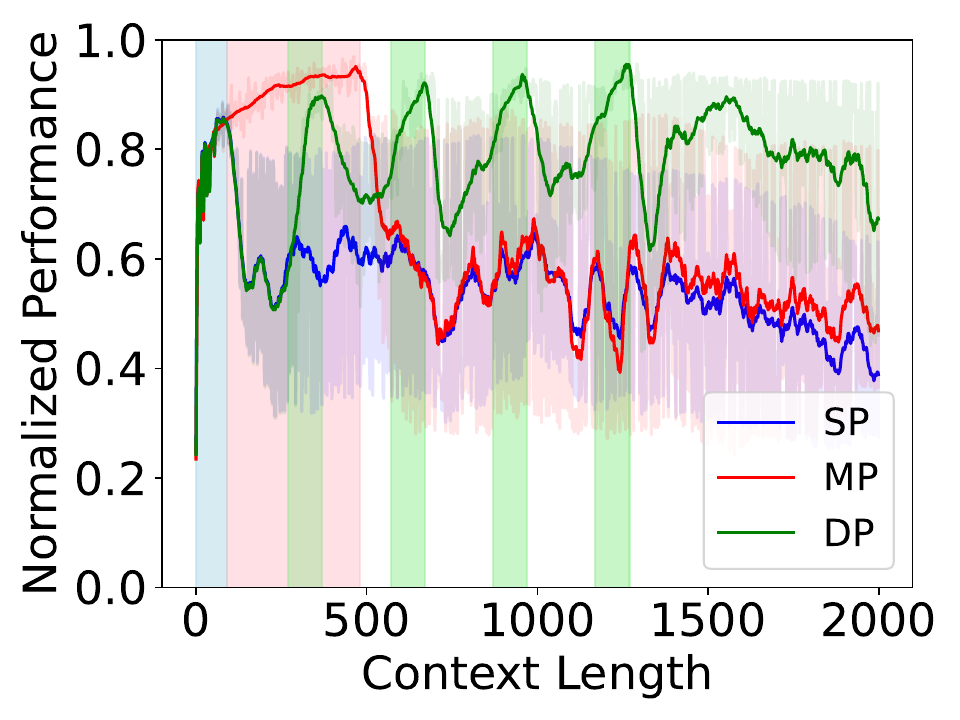}
    \caption{DEEPSEEK-R1}
    \label{fig:r1c1}
  \end{subfigure}\hfill
  \begin{subfigure}[b]{0.33\linewidth}
    \centering
    \includegraphics[width=\linewidth]{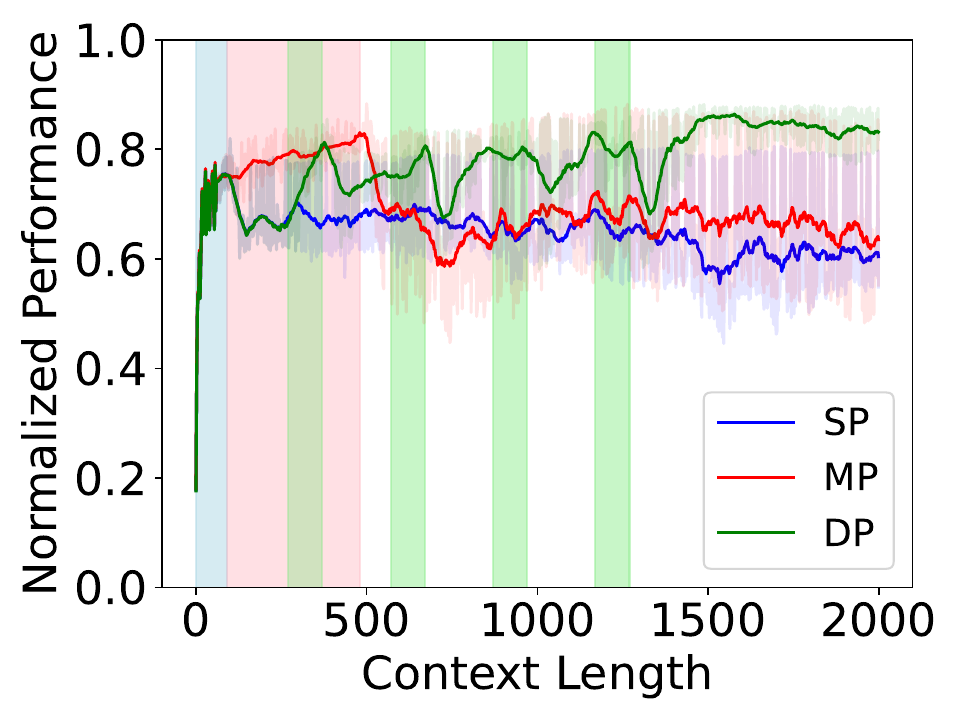}
    \caption{LLAMA3-8B}
    \label{fig:r1c2}
  \end{subfigure}\hfill
  \begin{subfigure}[b]{0.33\linewidth}
    \centering
    \includegraphics[width=\linewidth]{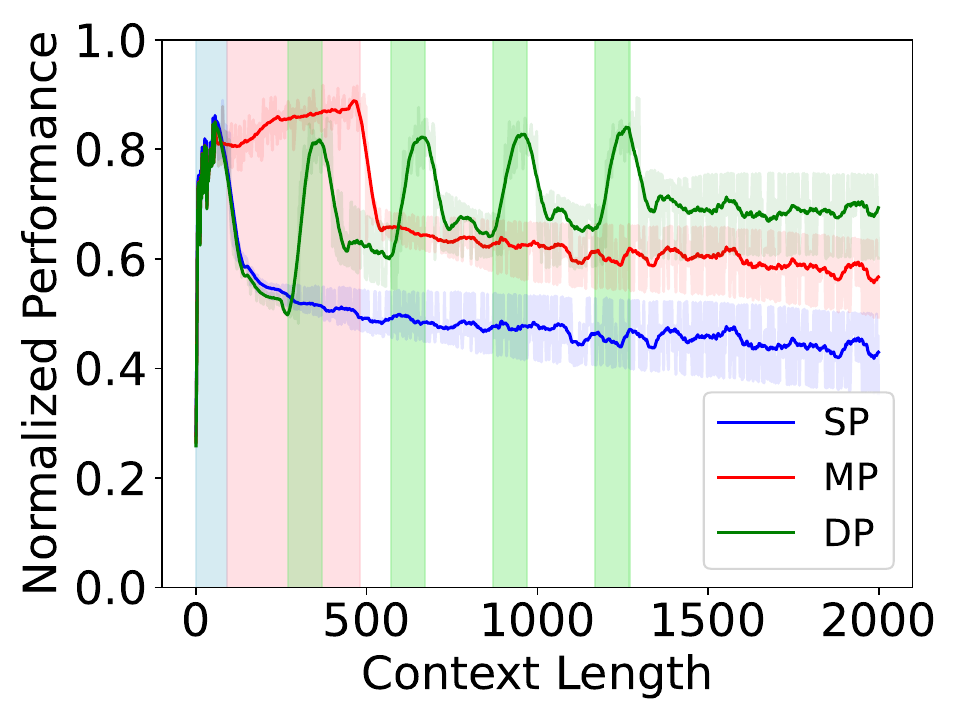}
    \caption{RWKV-7}
    \label{fig:r1c3}
  \end{subfigure}

  \vspace{0.6em}

  % ---------- Row 2 (3 figures) ----------
  \begin{subfigure}[b]{0.33\linewidth}
    \centering
    \includegraphics[width=\linewidth]{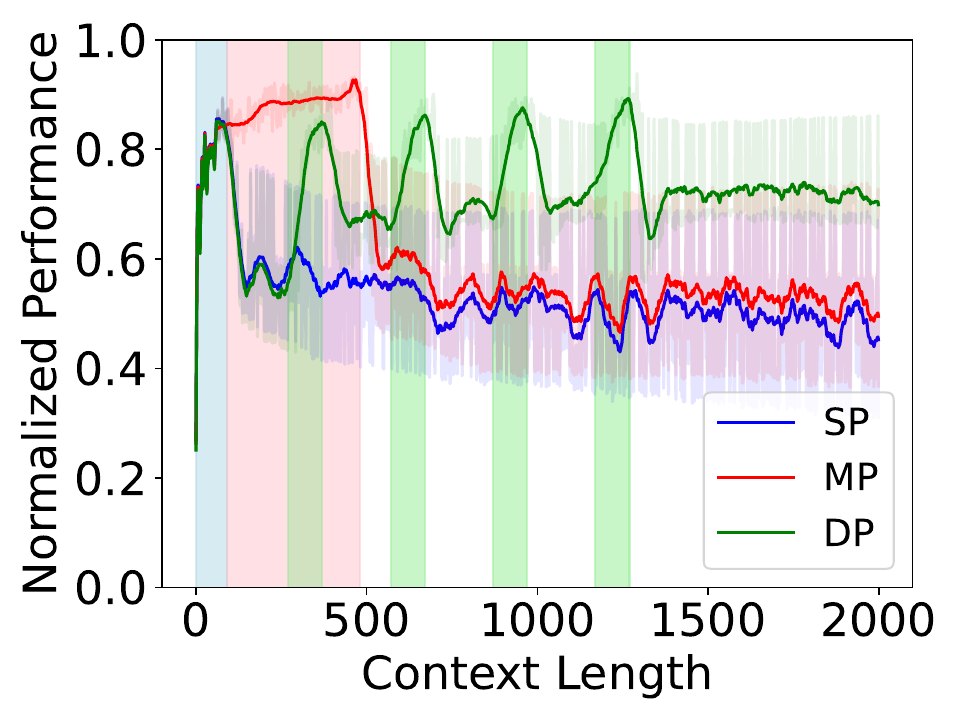}
    \caption{MAMBA}
    \label{fig:r2c1}
  \end{subfigure}\hfill
  \begin{subfigure}[b]{0.33\linewidth}
    \centering
    \includegraphics[width=\linewidth]{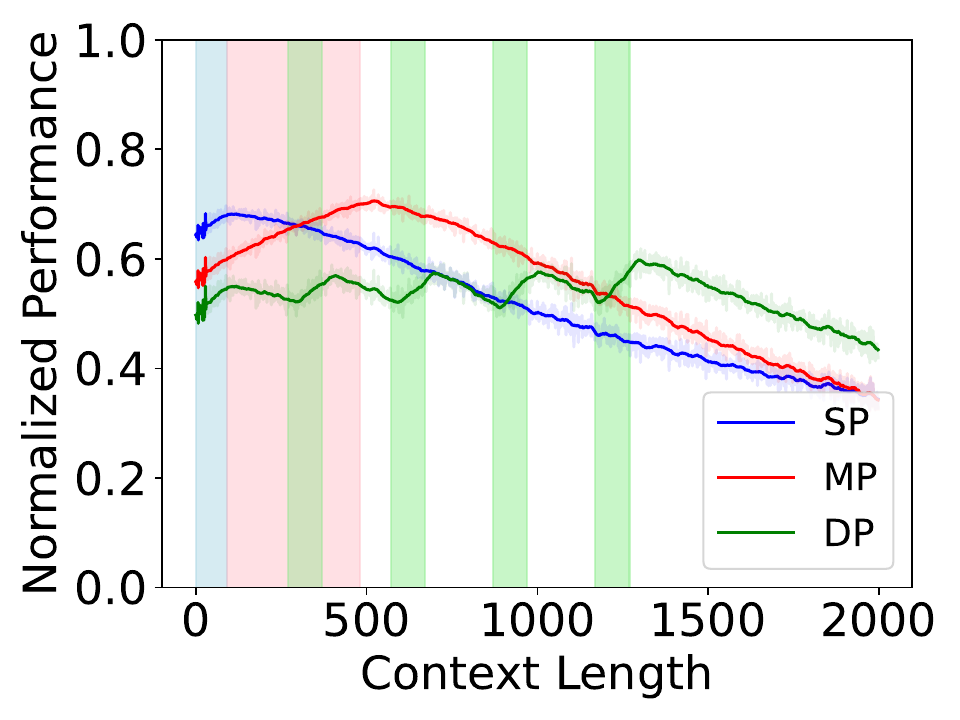}
    \caption{SGD}
    \label{fig:r2c2}
  \end{subfigure}\hfill
  \begin{subfigure}[b]{0.33\linewidth}
    \centering
    \includegraphics[width=\linewidth]{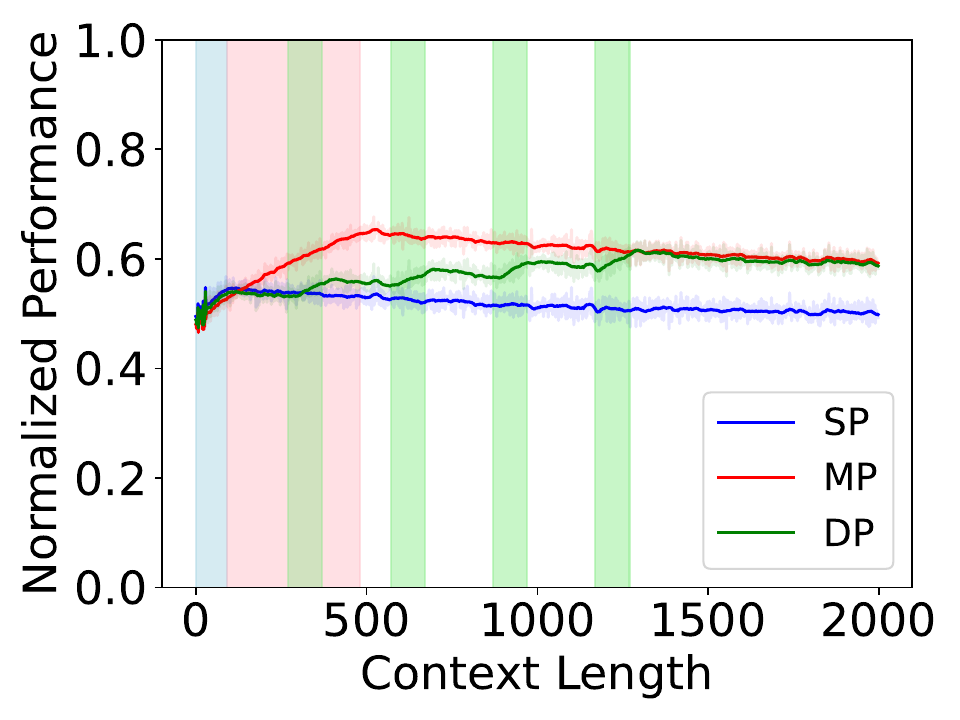}
    \caption{EWC}
    \label{fig:r2c3}
  \end{subfigure}

  \caption{Retention performance of ICCL and GBCL models on the simple task under SP, MP, and DP settings (with task identifiers, $\varphi_I=200$). Target task blocks are highlighted in light blue (SP), pink (MP), and green (DP).}
  \label{fig:simple_task_curves}
\end{figure}

\begin{figure}[h]
  \centering

  % ---------- Simple task ER ----------
  \begin{subfigure}[b]{0.33\linewidth}
    \centering
    \includegraphics[width=\linewidth]{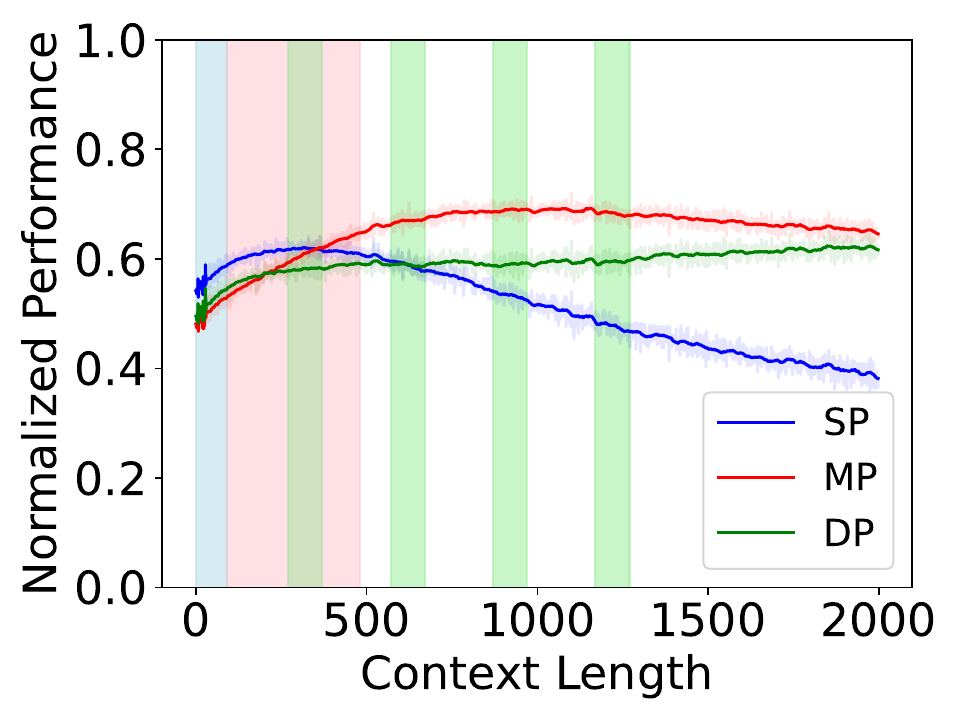}
    \caption{ER on simple task}
    \label{fig:er_simple}
  \end{subfigure}
  \hspace{0.02\linewidth}
  % ---------- Complex task ER ----------
  \begin{subfigure}[b]{0.33\linewidth}
    \centering
    \includegraphics[width=\linewidth]{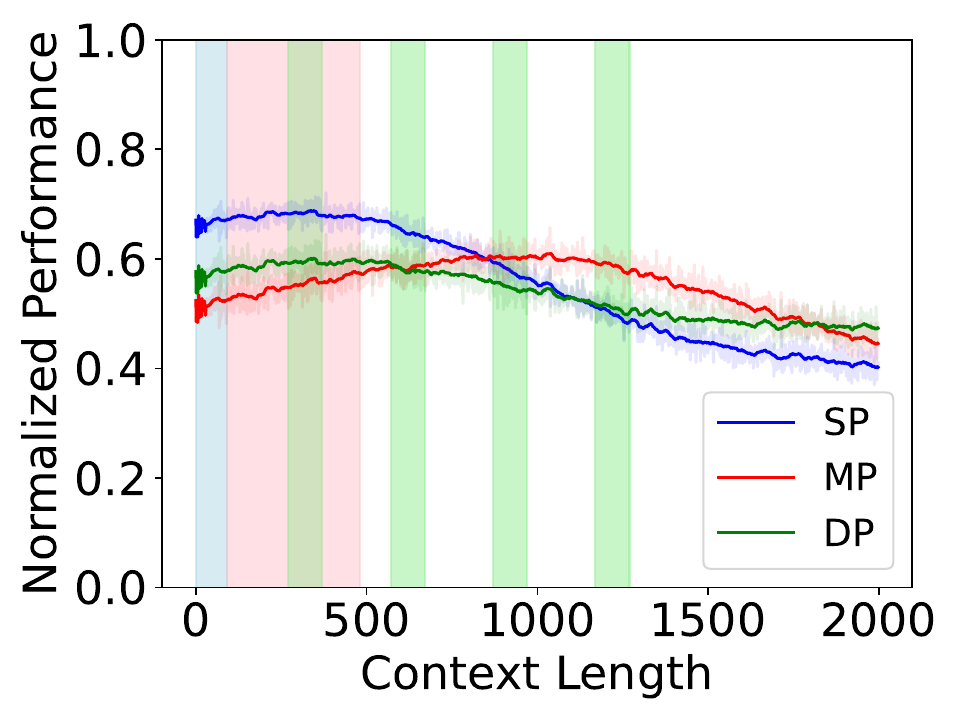}
    \caption{ER on complex task}
    \label{fig:er_complex}
  \end{subfigure}

  \caption{Retention performance of the ER baseline under SP, MP, and DP settings (with task identifiers, $\varphi_I=200$). Results are shown for both simple ($N_s=4$) and complex ($N_s=8$) Markov-chain tasks.}
  \label{fig:er_results}
\end{figure}

Figure~\ref{fig:simple_task_curves} illustrates the retention performance of ICCL models and GBCL baselines varies on the simple task with task identifiers and $\varphi_I=200$, under SP, MP, and DP scheduling. For ICCL methods, we observe that in both SP and MP conditions, the retention performance drops sharply immediately after finishing all target task demonstrations, and subsequently remains at a reduced level as the sequence length increases. In contrast, ICCL under DP consistently achieves better performance when the sequence length reaches its maximum, demonstrating its advantage in retaining cross-task knowledge. 
However, for GBCL baselines, DP does not yield a comparable improvement and performance remains unstable across sequence lengths. Detailed results for the ER baseline under simple and complex tasks are reported in Figure~\ref{fig:er_results}. These results confirm that both ICCL and GBCL models suffer from catastrophic forgetting, yet DP is effective in mitigating this issue for ICCL by alleviating performance degradation over long sequences.

\subsection{Optimal Parameter $\lambda$ in EWC}

\begin{figure}[h]
    \centering
    \begin{subfigure}[b]{0.3\textwidth}
        \centering
        \includegraphics[width=\linewidth]{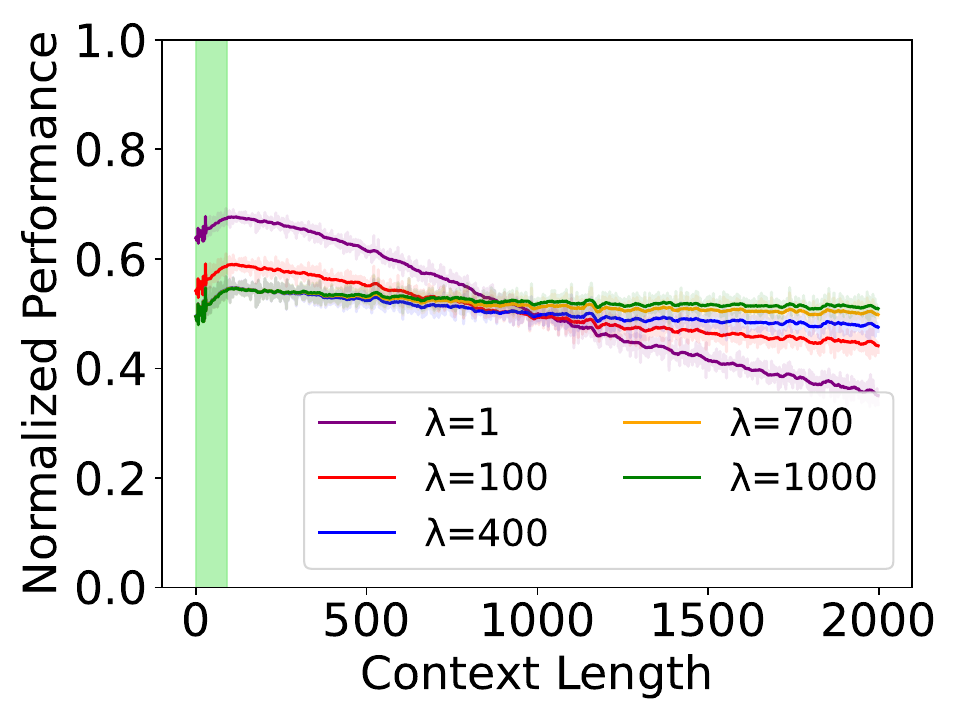}
        \caption{Performances under SP}
        \label{fig:model1}
    \end{subfigure}
    \hfill
    \begin{subfigure}[b]{0.3\textwidth}
        \centering
        \includegraphics[width=\linewidth]{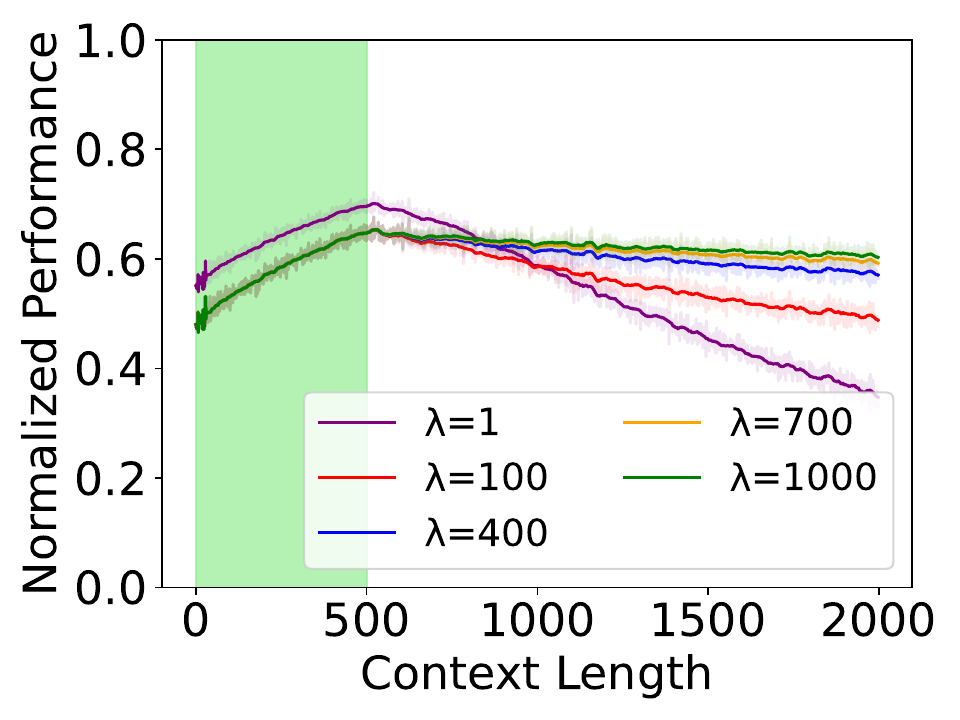}
        \caption{Performances under MP}
        \label{fig:model2}
    \end{subfigure}
    \hfill
    \begin{subfigure}[b]{0.3\textwidth}
        \centering
        \includegraphics[width=\linewidth]{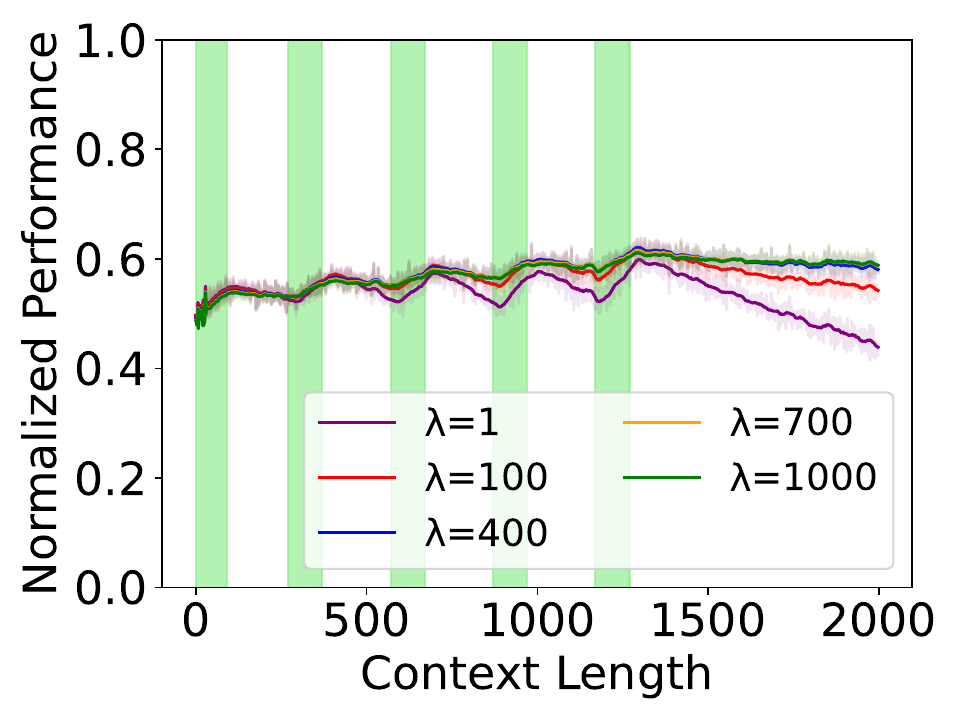}
        \caption{Performances under DP}
        \label{fig:model3}
    \end{subfigure}

    \caption{EWC's retention performances under Markov chains with simple task (markov chain with $Ns$ = 4).}
    \label{fig:ewc_lambda_simple}
\end{figure}

\begin{figure}[ht]
    \centering
    \begin{subfigure}[b]{0.3\textwidth}
        \centering
        \includegraphics[width=\linewidth]{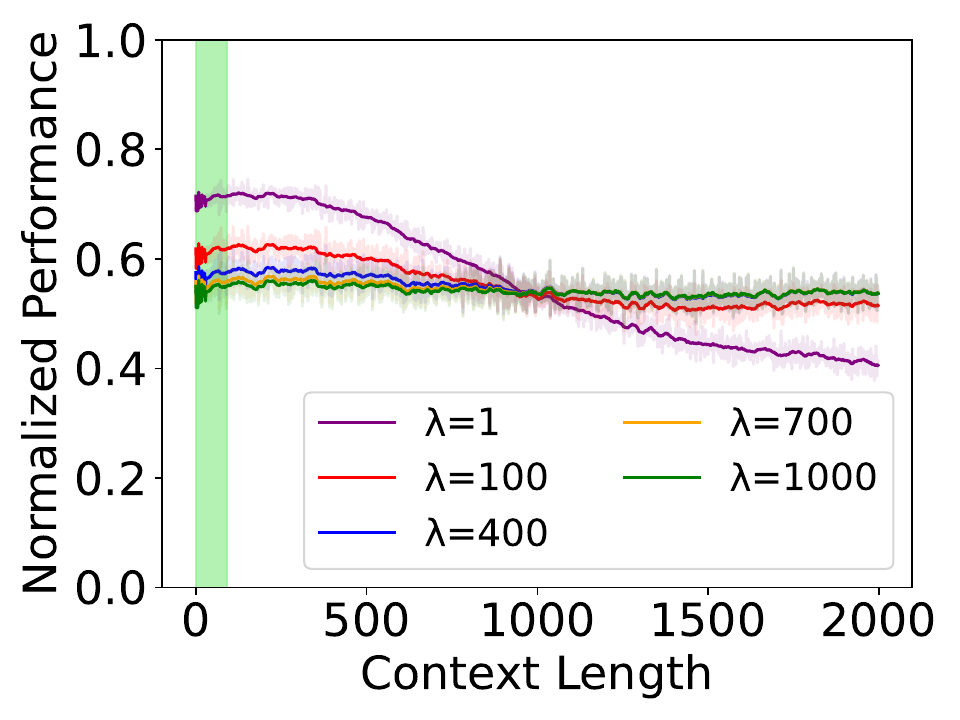}
        \caption{Performances under SP}
        \label{fig:model1}
    \end{subfigure}
    \hfill
    \begin{subfigure}[b]{0.3\textwidth}
        \centering
        \includegraphics[width=\linewidth]{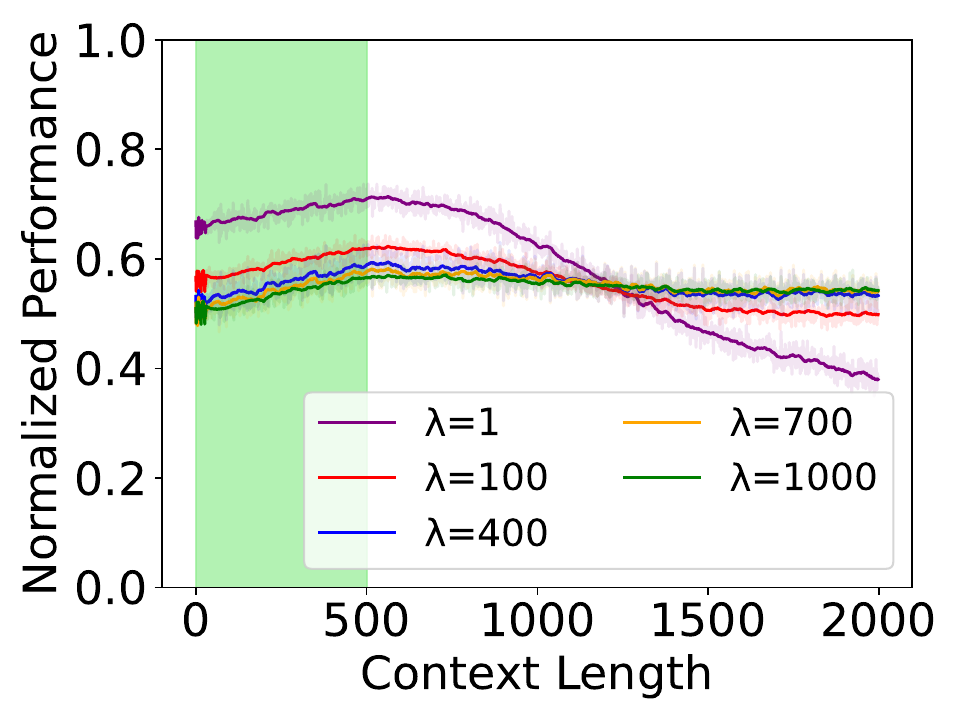}
        \caption{Performances under MP}
        \label{fig:model2}
    \end{subfigure}
    \hfill
    \begin{subfigure}[b]{0.3\textwidth}
        \centering
        \includegraphics[width=\linewidth]{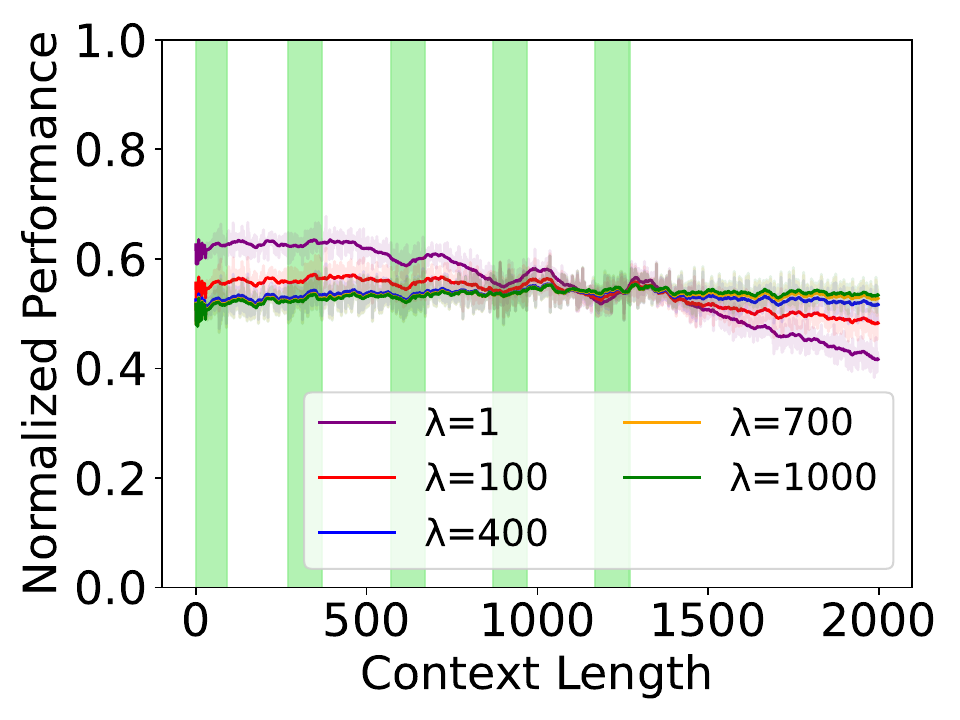}
        \caption{Performances under DP}
        \label{fig:model3}
    \end{subfigure}

    \caption{EWC's retention performances under Markov chains with complex task (markov chain with $Ns$ = 8).}
    \label{fig:ewc_lambda_complex}
\end{figure}

Figure~\ref{fig:ewc_lambda_simple} and Figure~\ref{fig:ewc_lambda_complex} present the performance of EWC under simple and complex tasks with different regularization strengths $\lambda_{\text{EWC}}$ across SP, MP, and DP scheduling strategies. The results highlight the trade-off between plasticity and stability inherent in EWC. With small $\lambda_{\text{EWC}}$ values (e.g., $\lambda_{\text{EWC}}=100$), the model maintains higher plasticity and adapts quickly to new demonstrations; however, this comes at the cost of poor stability, as retention performance decays sharply with increasing context length. Conversely, large $\lambda_{\text{EWC}}$ values (e.g., $\lambda_{\text{EWC}}=1000$) enforce stronger stability but severely limit plasticity, leading to under-adaptation and overall lower performance across all scheduling strategies.

Intermediate values achieve a better balance. In particular, $\lambda_{\text{EWC}}=700$ consistently provides a favorable trade-off, delivering higher long-term retention compared to smaller $\lambda_{\text{EWC}}$ while avoiding the rigidity observed at $\lambda_{\text{EWC}}=1000$. This effect is especially evident under MP and DP, where $\lambda_{\text{EWC}}=700$ maintains relatively stable retention curves while smaller $\lambda_{\text{EWC}}$ values degrade more rapidly. Based on these observations, we select $\lambda_{\text{EWC}}=700$ as the default regularization strength for subsequent experiments, as it offers the most robust balance between plasticity and stability across both simple and complex tasks

\begin{table}[t]
\caption{The comparisons between the true normalized performance of ICCL and GBCL methods and their fitted ACT-R curves.}
\label{tab:fitted_values}
% \centering
\resizebox{\linewidth}{!}{
\begin{tabular}{lccccccc}
\hline
\textbf{Model} & DEEPSEEK-R1 & LLAMA3-8B & RWKV-7 & MAMBA & SGD & ER & EWC\\
\hline
Pearson \\ correlation  & 0.5152 & 0.7068 & 0.6171 & 0.6215 & 0.0339 & 0.5660 & 0.7253 \\
\hline
\end{tabular}
}
\end{table}

\subsection{The Fitted ACT-R model results from ICCL and GBCL}
\begin{figure}[htp]
    \centering
    \includegraphics[width=0.9\linewidth]{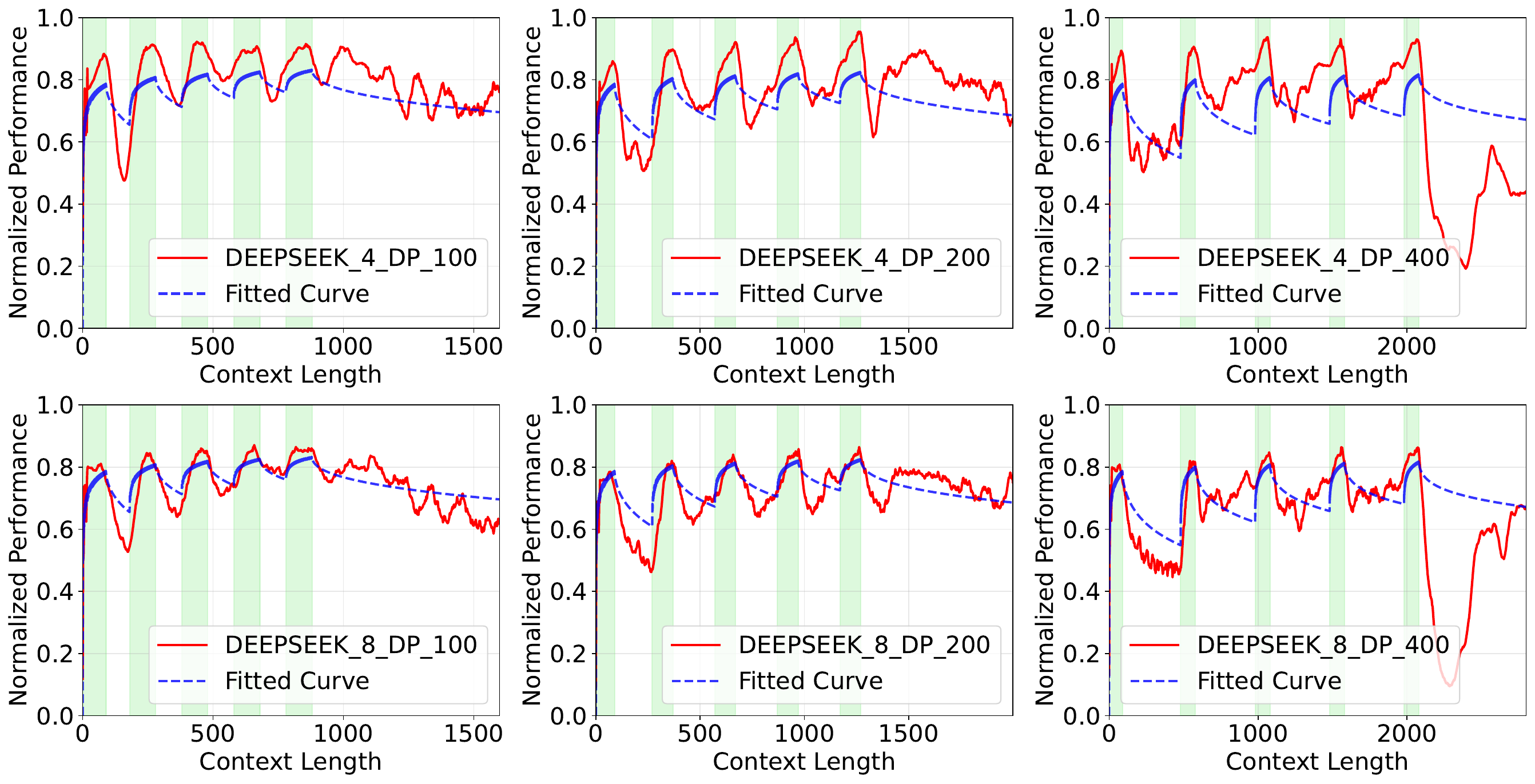}
    \caption{The performance comparisons between DEEPSEEK-R1 and its fitted ACT-R model.}
    \label{fig:similarity_deepseek}
\end{figure}

\begin{figure}[htp]
    \centering
    \includegraphics[width=0.9\linewidth]{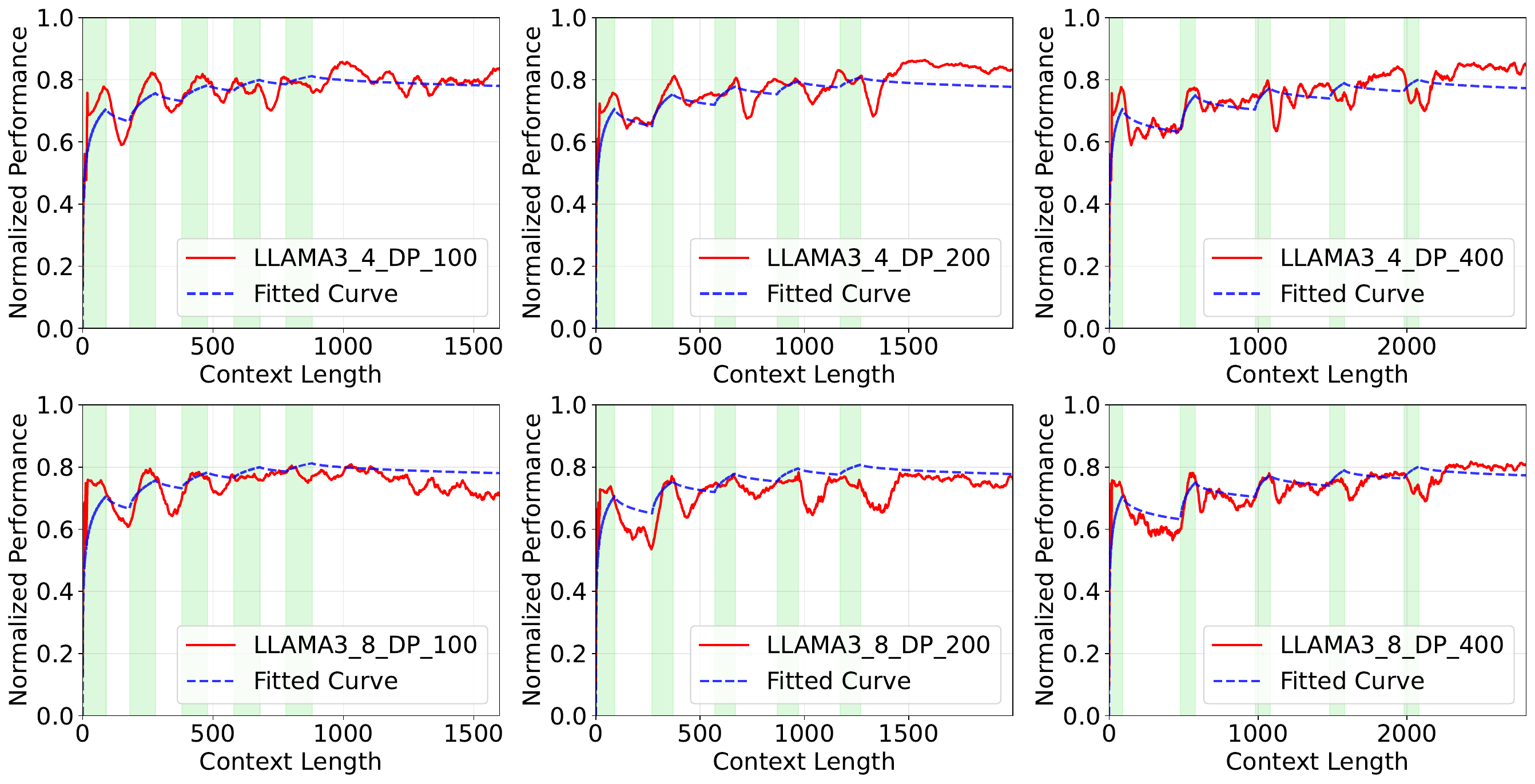}
    \caption{The performance comparisons between LLAMA3-8B and its fitted ACT-R model.}
    \label{fig:similarity_llama}
\end{figure}

\begin{figure}[htp]
    \centering
    \includegraphics[width=0.9\linewidth]{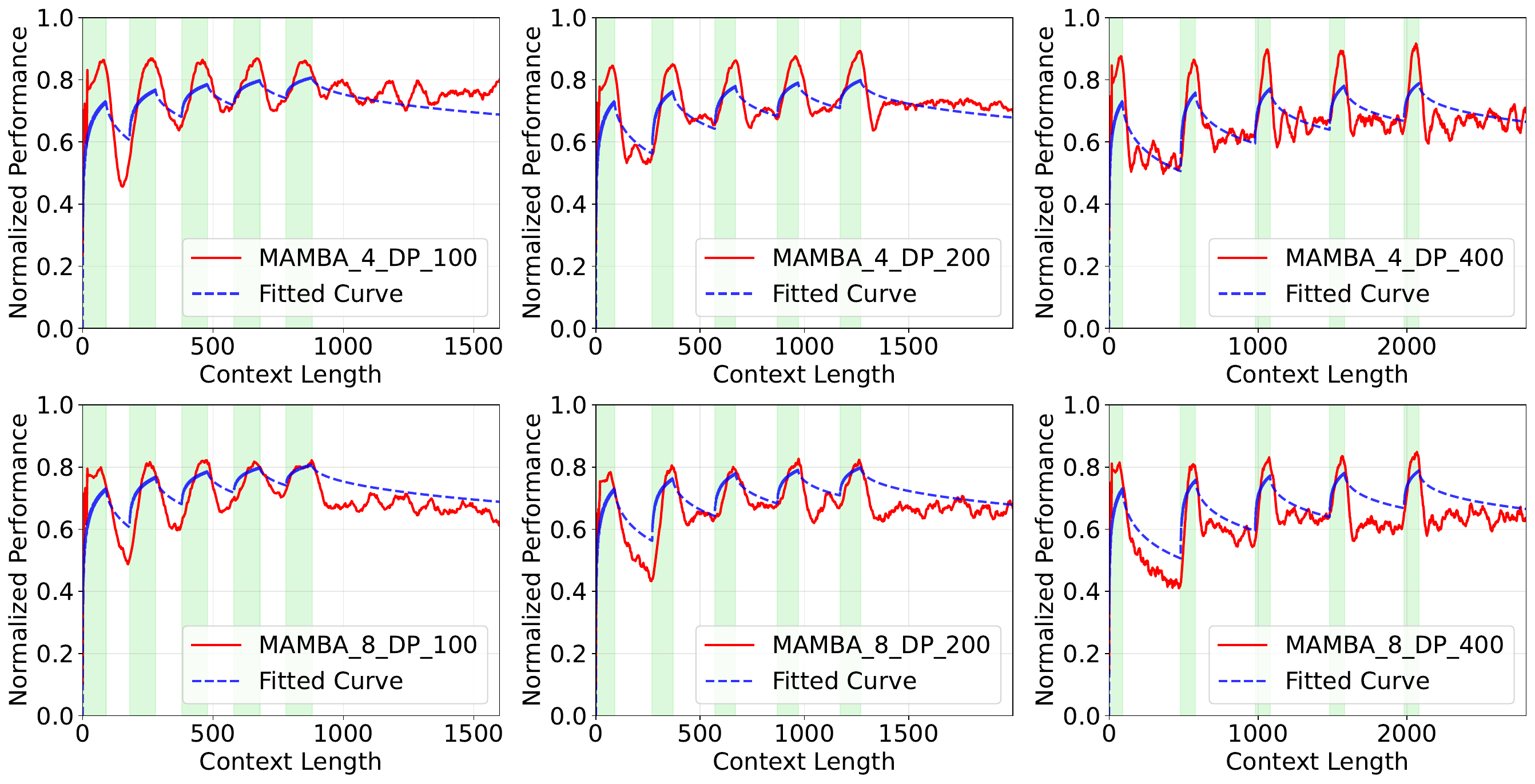}
    \caption{The performance comparisons between MAMBA and its fitted ACT-R model.}
    \label{fig:similarity_mamba}
\end{figure}

\begin{figure}[htp]
    \centering
    \includegraphics[width=0.9\linewidth]{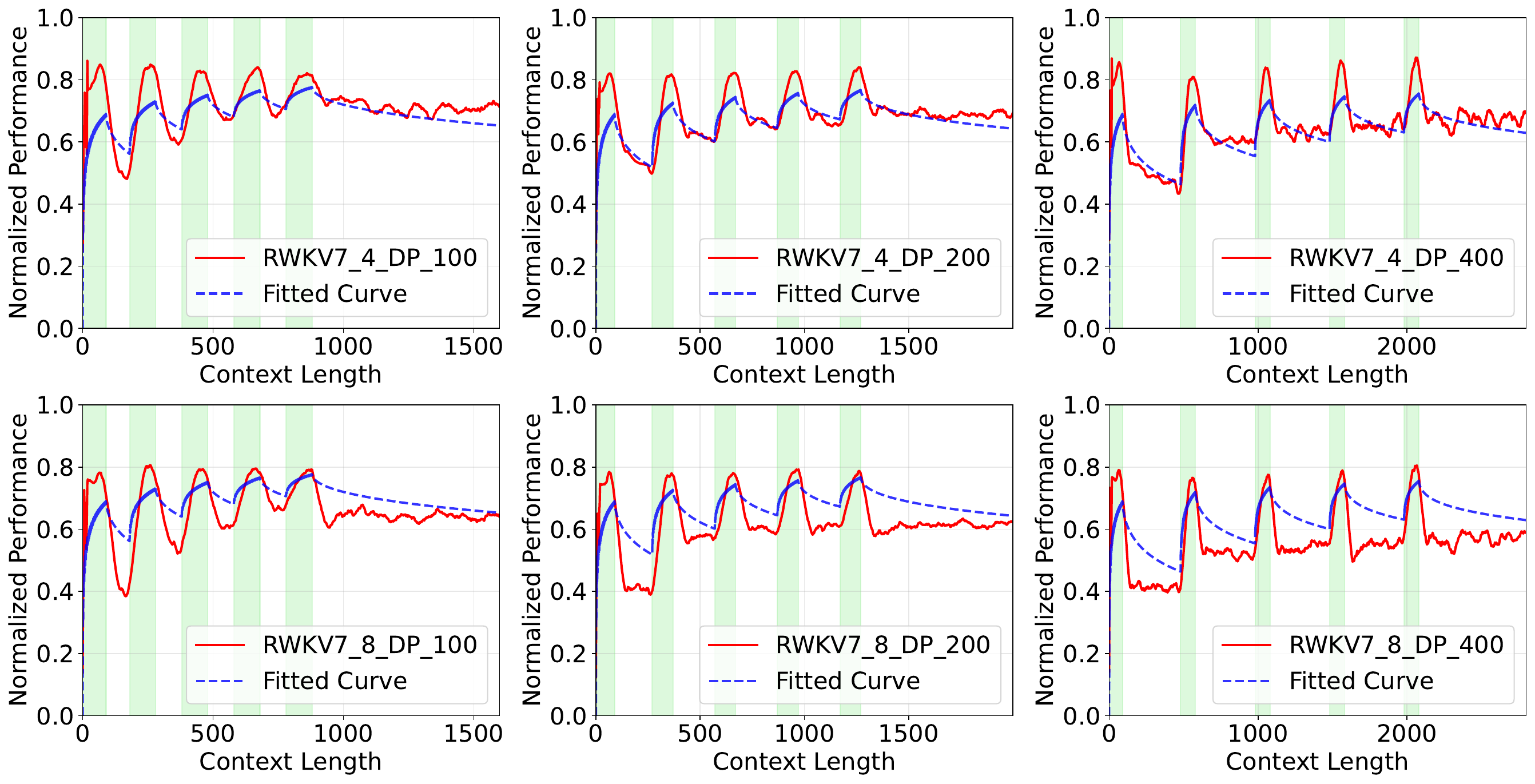}
    \caption{The performance comparisons between RWKV-7 and its fitted ACT-R model.}
    \label{fig:similarity_rwkv}
\end{figure}

\begin{figure}[htp]
    \centering
    \includegraphics[width=0.9\linewidth]{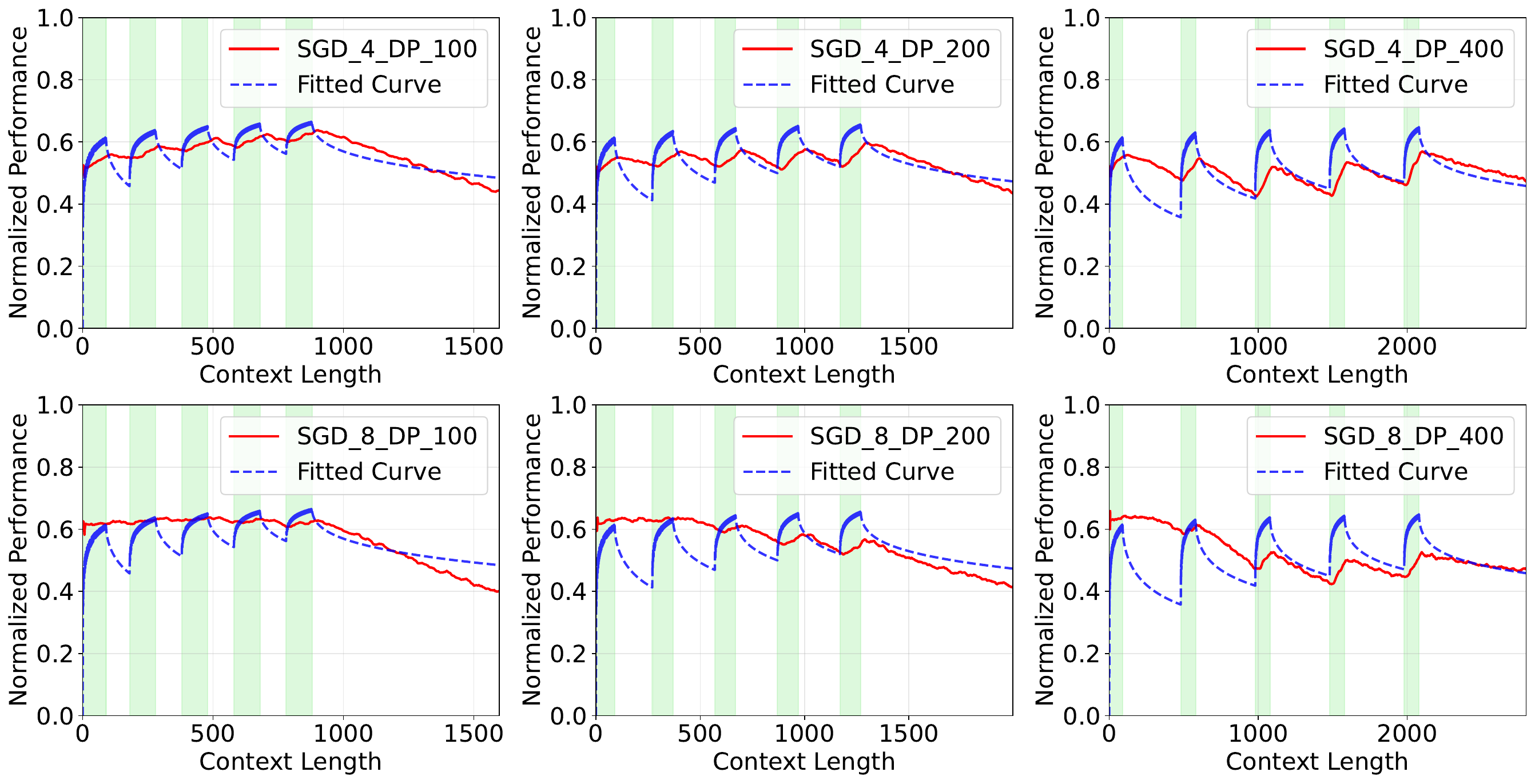}
    \caption{The performance comparisons between SGD and its fitted ACT-R model.}
    \label{fig:similarity_sgd}
\end{figure}

\begin{figure}[htp]
    \centering
    \includegraphics[width=0.9\linewidth]{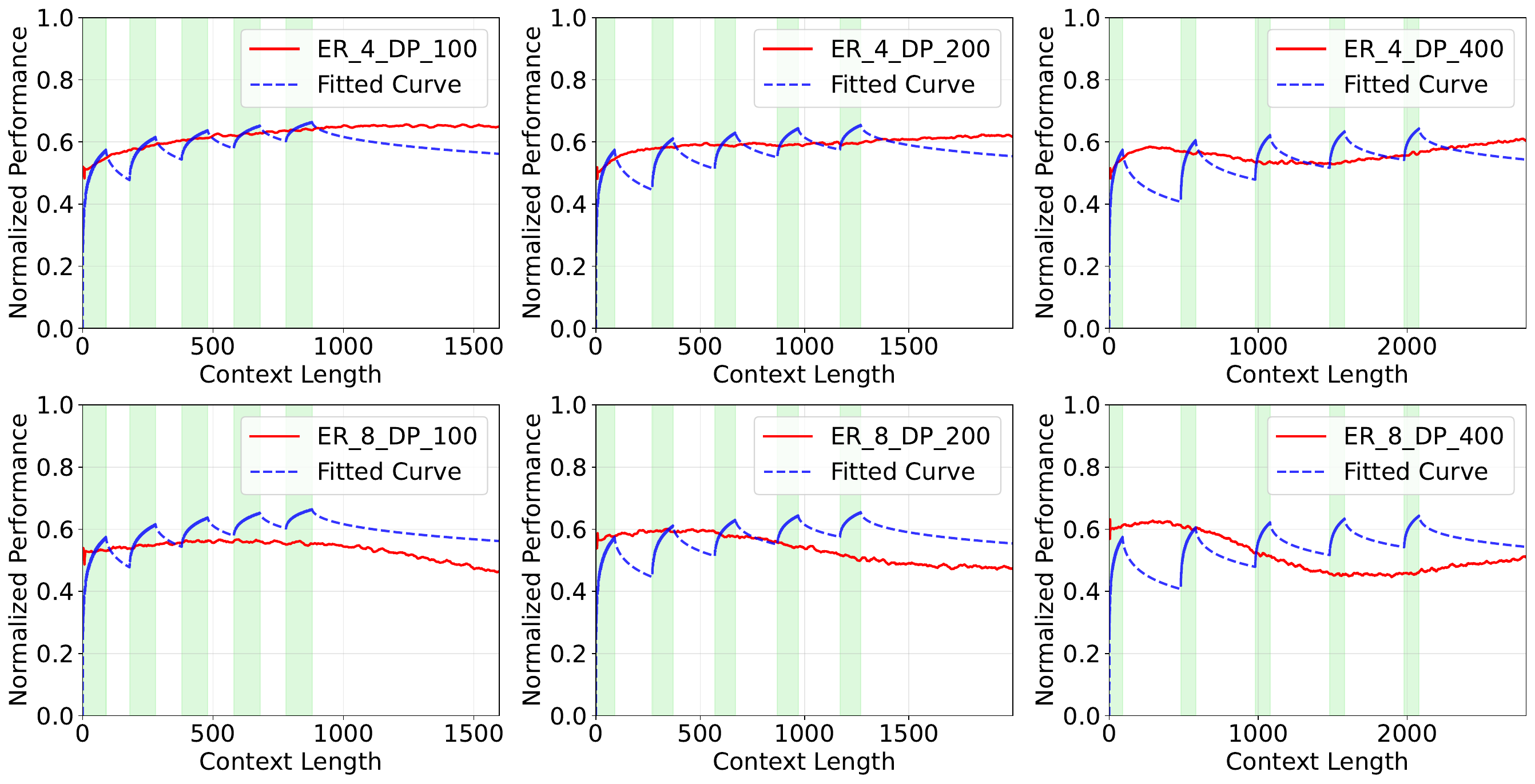}
    \caption{The performance comparisons between ER and its fitted ACT-R model.}
    \label{fig:similarity_er}
\end{figure}

\begin{figure}[p]
  \centering
  \includegraphics[width=0.9\linewidth]{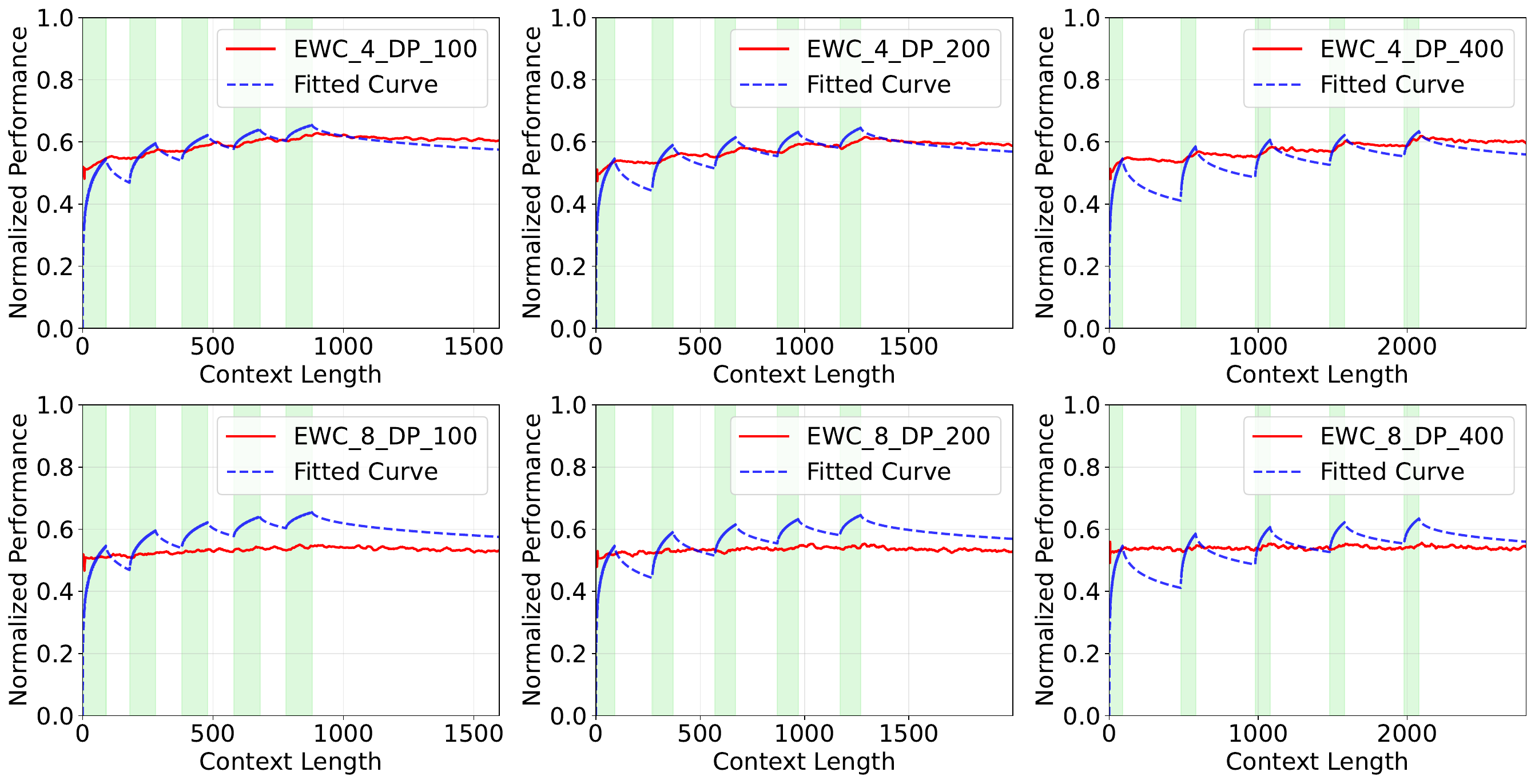}
  \caption{The performance comparisons between EWC and its fitted ACT-R model.}
  \label{fig:similarity_ewc}
\end{figure}

To further analyze the retention dynamics of ICCL and GBCL methods, we leveraged the experimental data of normalized retention performance across prompt sequence positions under the DP setting. We applied a modified ACT-R model to fit the performance trajectories of each method. The comparisons between the observed normalized retention performance curves and the corresponding fitted ACT-R curves on both simple and complex tasks are illustrated in Figures~\ref{fig:similarity_deepseek}--\ref{fig:similarity_ewc}. 

In addition, we computed the average Pearson correlation coefficients between the true normalized retention performance curves and their fitted ACT-R counterparts, as reported in Table~\ref{tab:fitted_values}. The results show that, with the exception of SGD, all ICCL and GBCL methods achieve relatively high correlation coefficients (e.g., 0.71 for LLaMA3-8B and 0.73 for EWC). This indicates that the fitted ACT-R model provides a reliable description of the dynamics of normalized retention performance, capturing how ICCL and GBCL retention evolves with increasing sequence positions.

% \begin{figure}[t]
%     \centering
%     \includegraphics[width=0.9\linewidth]{figures/ewc_similarity_final.pdf}
%     \caption{The normalized performance comparisons between EWC and its fitted ACT-R model.}
%     \label{fig:similarity_ewc}
% \end{figure}

\makeatletter
\setlength{\@fptop}{0pt}
\makeatother

\end{document}